\documentclass[journal]{IEEEtran}
\ifCLASSINFOpdf
\else
\fi

\usepackage{graphicx}
\usepackage[linesnumbered,ruled,lined]{algorithm2e}
\usepackage{amssymb}
\usepackage{amsmath}
\usepackage[left,pagewise]{lineno}
\usepackage{colortbl,booktabs}
\usepackage{threeparttable}
\usepackage{setspace}
\usepackage{bm}
\usepackage{url}
\usepackage{subfigure}

\usepackage{xcolor}
\usepackage{multirow}

\begin{document}

%
%
\title{Masked Spatial-Spectral Autoencoders Are Excellent Hyperspectral Defenders} 

\author{\IEEEauthorblockN{Jiahao~Qi, Zhiqiang~Gong, Xingyue~Liu, Kangcheng~Bin, Chen~Chen, Yongqian~Li, Wei~Xue, Yu~Zhang, and Ping~Zhong~\IEEEmembership{Senior Member,~IEEE}}
\thanks{This work was supported in part by the Natural Science Foundation of China under Grant 61971428 and Grant 62001502. \emph{(Corresponding author: Ping Zhong.)}}
\thanks{Jiahao Qi, Xingyue Liu, and Ping Zhong are with the National Key Laboratory of Science and Technology on Automatic Target Recognition, National University of Defense Technology, Changsha 410073, China (e-mail: qijiahao1996@nudt.edu.cn, xingyueliu0801@nudt.edu.cn, binkc21@nudt.edu.cn, chenchen21c@nudt.edu.cn, liyongqian@nudt.edu.cn, zhangyu13a@nudt.edu.cn, zhongping@nudt.edu.cn).}
\thanks{Zhiqiang Gong is with the National Innovation Institute of Defense Technology, Chinese Academy of Military Science, Beijing 100000, China (e-mail: gongzhiqiang13@nudt.edu.cn).}
\thanks{Wei Xue is with the School of Computer Science and Technology, Anhui University of Technology, Maanshan 243032, China (e-mail: cswxue@ahut.edu.cn).}
}

\markboth{Submitted to IEEE Transactions on Image Processing}%
{Qi \MakeLowercase{\textit{et al.}}: Masked Spatial-Spectral Autoencoder Are Excellent Hyperspectral Defender}
%



\IEEEtitleabstractindextext{%
\begin{abstract}
  Deep learning methodology contributes a lot to the development of hyperspectral image (HSI) analysis community. 
  However, it also makes HSI analysis systems vulnerable to adversarial attacks. 
  To this end, we propose a masked spatial-spectral autoencoder (MSSA) in this paper under self-supervised learning theory, for enhancing the robustness of HSI analysis systems. 
  First, a masked sequence attention learning module is conducted to promote the inherent robustness of HSI analysis systems along spectral channel. 
  Then, we develop a graph convolutional network with learnable graph structure to establish global pixel-wise combinations.
  In this way, the attack effect would be dispersed by all the related pixels among each combination, and a better defense performance is achievable in spatial aspect.
  Finally, to improve the defense transferability and address the problem of limited labelled samples, MSSA employs spectra reconstruction as a pretext task and fits the datasets in a self-supervised manner.
  Comprehensive experiments over three benchmarks verify the effectiveness of MSSA in comparison with the state-of-the-art hyperspectral classification methods and representative adversarial defense strategies. 
\end{abstract}

\begin{IEEEkeywords}
Deep learning, Hyperspectral image analysis, Adversrarial defense, Input masking, Graph convolutional network, Self-supervised learning, Spectral reconstruction.
\end{IEEEkeywords}}

\maketitle

\IEEEdisplaynontitleabstractindextext

%
\IEEEpeerreviewmaketitle

\section{Introduction}\label{sec:1}

\IEEEPARstart{I}{n} recent years, hyperspectral image (HSI) with high spatial resolution and hundreds of spectral channels, has attracted increasing attention in image analysis community.
Aided by the compelling development of deep learning (DL) methodology, HSI analysis achieves the state-of-the-art performances in many high-level applications, such as scenario classification \cite{Dong2022,Wu2018}, target detection \cite{Yang2016,Wang2017}, and image super-resolution \cite{DongW2022, Zhu2021}. 
\par
\begin{figure}[t]
  \centering
    \includegraphics[scale=0.55]{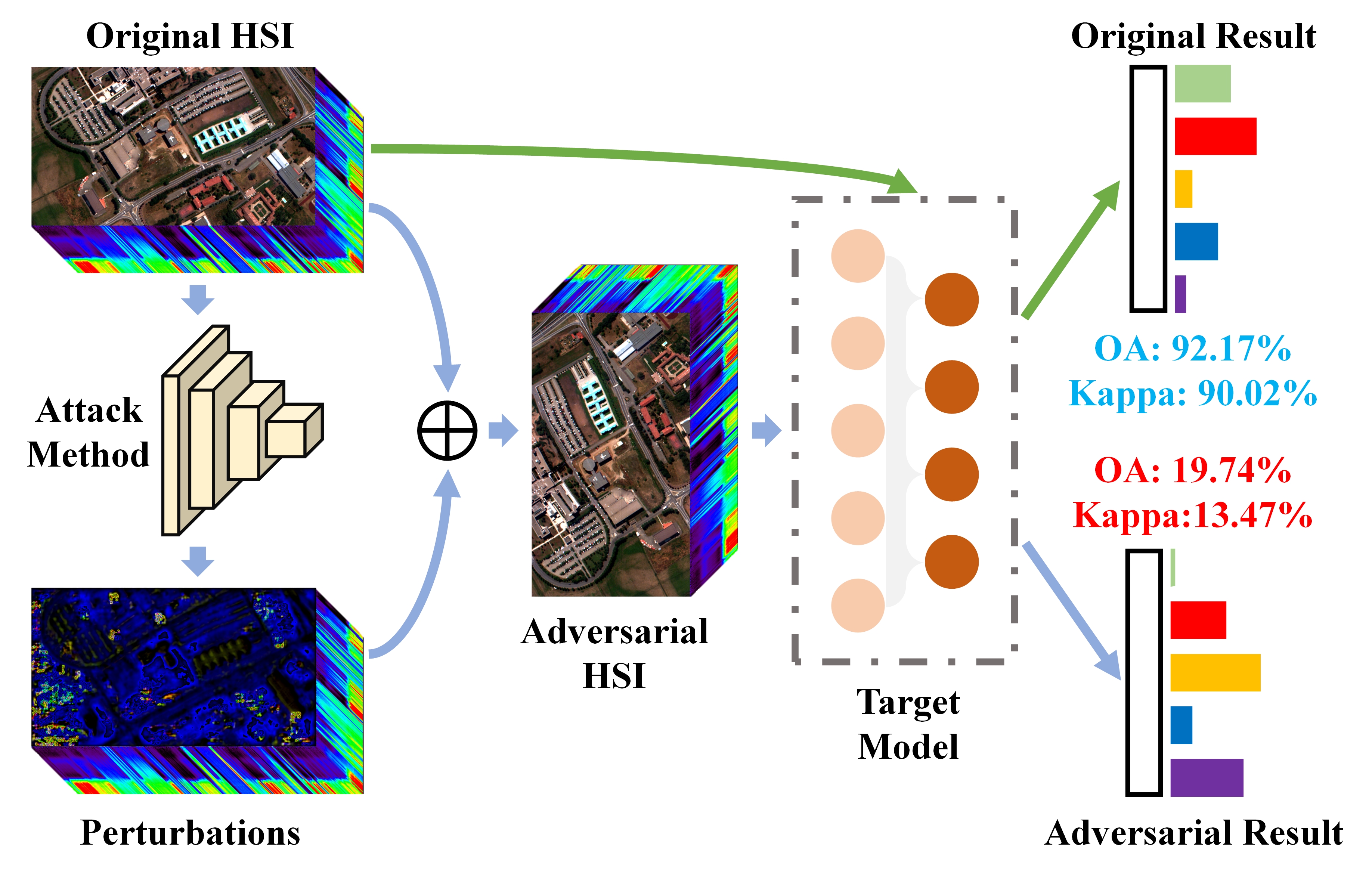}
     \caption{The process of using adversarial examples to attack DL based methods, taking hyperspectral image classification task as an instance.}
  \label{fig:A1}
  \end{figure}

However, prior works \cite{Xu2021,Park2021} confirms that, the DL-based HSI analysis methods are vulnerable to adversarial examples, which are generated by inflicting imperceptible modifications on input datasets.
As shown in Fig. \ref{fig:A1}, a dramatic performance drop of DL model occurs after being attacked by adversarial example. 
Under this circumstance, the attacked DL model performs poorly or even generates misleading results.
Consequently, it is necessary to attach importance to the development of hyperspectral adversarial defense strategies, particularly in the risk-critical scenarios.
\par 
Defending against attacks emerges as a dominant research topic in image processing field and numerous approaches have been proposed in the past decade.
These methods could be loosely split into three categories \cite{Akhtar2021}: adversarial training, adversarial detection, and input transformation.
Adversarial training is the most commonly used defense strategy owing to its easy implementation.
It proactively exposes a DL model to the generated adversarial examples during the training stage for building immunity towards adversarial attacks, by utilizing different adversarial example generation tricks \cite{Madry2017,Lin2020,Karras2021}.
Instead of establishing attack immunity in training stage, adversarial detection works as an add-on mechanism or module to detect adversarial samples and refine input datasets at the inference phase.
As a representative, Qin \emph{et al}. \cite{Qin2020} devised a special mechanism known as class-conditional reconstruction to solve adversarial example detection problems.
In contrast to the previous two defense measures, input transformation focuses on enhancing the robustness of DL models by improving data security. 
It alters the data structures of input samples with some strategie, so that the modified samples can become more stable when being attacked. 
Among approaches of this class, image encoding, such as thermometer encoding method \cite{Buckman2018}, is the representative one to resist the adversarial perturbations from inputs. 
\par
There is no doubt that remarkable success has been achieved in adversarial defense community. 
Unfortunately, the majority of existing adversarial defense strategies are mainly developed for RGB domain \cite{Xu2021}. 
Even worse, these methods can not be directly applied to hyperspectral domain for the following reasons:
\par
\textbf{(\romannumeral1) Limited labelled samples:} Unlike traditional RGB training datasets, such as ImageNet \cite{Deng2009}, which contain tens of millions of labelled samples, hyperspectral datasets have such a small number of labelled samples that they can not even characterize the statistical features of each category.
Under this circumstance, generation-based defense methods like adversarial training may not perform well.
Because it is not trivial to yield approving adversarial examples with limited labelled training data. 
\par
\textbf{(\romannumeral2) Poor transfer capability:} Most of existing defense mechanisms are especially designed to counter some specific adversarial attacks \cite{Akhtar2021}.
Taking adversarial detection as an example, if the training adversarial samples are generated under a $l_{2}$-norm bounded constraint, the adversarial detection methods are not able to detect $l_{p}$-norm bounded perturbations (when $p \neq 2$ ) \cite{Zhang2019}. 
In other words, conventional defense methods do not promote the intrinsic resistibility of DL models and then possess an insufficient transfer capability. 
\par
\textbf{(\romannumeral3) Heavy computation burden:} The enormous spectral bands of HSI contain a large amount of redundant information, which result in Hughes phenomenon \cite{Hughes1968} and more computation consumption. 
Whereas, several input transformation methods, such as input encoding \cite{Buckman2018}, would further increase the volume of input HSI. 
As a result, these methods cannot be applied in real-world applications directly, owing to the heavy computation burden.
\par
Facing with the aforementioned problems, it is challenging and crucial to investigate sample-efficient, transferable, and lightweight hyperspectral adversarial defense techniques.
In this paper, we present a novel and effective network, named masked spatial-spectral autoencoder (MSSA), to enhance the inherent robustness of HSI analysis systems. 
MSSA conducts adversarial defense in both spectral and spatial aspects.
More specifically, MSSA develops a one-dimensional (1-D) random masking mechanism to handle input HSI along the spectral channel, with only unmasked spectral patches being utilized as the input of encoder network.
In this way, most adversarial perturbations are distilled before implementing attack, while the inconsecutive forms would further reduce the attack effect of these perturbations. 
Besides, this mechanism also helps to tackle the unbearable computation burden challenge, since only a few percentage of spectral information is utilized after the inputs are randomly masked. 
As for the spatial aspect, we design a dynamic graph convolutional network (GCN) block. 
With this block, the prediction of current pixel is based on not only its own characteristics but also those of its neighbors, while the attack effect would also be dispersed by these neighborhood pixels. 
Therefore, attack methods require a higher level of perturbations to successfully attack a DL network, resulting in a better adversarial defense performance in the spatial aspect. 
\par
Furthermore, MSSA performs adversarial defense without using any prior knowledge of the adversarial examples. 
In other words, MSSA is not designed for some specific attack methods and possesses a remarkable transfer capability. 
Moreover, the combination of input random masking mechanism and encoder-decoder architecture makes it possible to train MSSA in a self-supervised manner, which can effectively address the problem of limited labelled samples. 
The main contributions of this work can be briefly summarized as below.
\par
1) We propose a novel adversarial defense method named MSSA for HSI analysis systems. 
It devotes to enhance the inherent robustness of DL models. 
MSSA could be considered as an exceptional and robust feature extractor for most of downstream tasks (scenario classification, target detection, image super-resolution, etc.) in hyperspectral DL domain.
\par
2) A 1-D random masking mechanism are introduced to dispose the input spectra for easing the attack impact from spectral aspect, while spectrum reconstruction is specified as the pretext task. 
In this way, MSSA could fit the training datasets in a self-supervised fashion, which contributes a lot to solve the limited labelled samples problem and alleviate the heavy computation burden.
To the best of our knowledge, it is the first time to investigate the adversarial defense strategies with a self-supervised based input masking mechanism.
\par
3) In order to achieve a better adversarial defense effect from spatial aspect, we develop a special GCN block based on graph structure learning methodology. 
With the help of this GCN block, attack effect is shared among the connected graph nodes and MSSA is able to make more robust decisions. 
\par
4) Classification task is employed as an instance to conduct corresponding experiments for validating the efficiency and effectiveness of MSSA.
The corresponding results and analyses have clearly demonstrated that MSSA dramatically enhances the intrinsic resistibility of DL models and possesses a favorable transfer capability to a wider range of attack methods. 
\par
The remainder of this paper is organized as follows.
The motivations of developing our proposed method are introduced and discussed in Section \ref{sec:3}.
In Section \ref{sec:4}, we introduce the implementation details for the proposed method. 
Performance comparisons and their corresponding analyses are presented in Section \ref{sec:5}.
Finally, Section \ref{sec:6} makes a conclusion about the whole paper and looks forward to the future work.

\section{Motivations}\label{sec:3} 
In this section, we will briefly introduce the motivations adopted in this work for a better understanding of the proposed method. 
Specifically, MSSA enhances the inherent robustness of deep neural network towards adversarial attacks with two dominant principles: adversarial perturbation sparsification (spectral aspect) and global pixel-wise combination defense (spatial aspect). 
\par 


\subsection{Adversarial Perturbation Sparsification}
Input masking may not be a brand new strategy which was originally used for data augmentation in image classification tasks. 
However, it turns into a research focus since masked encoding methods are confirmed to learn more discriminative representations from the inputs corrupted by masking.
Then, numerous masked DL frameworks are proposed in both natural language processing (NLP) and computer vision (CV) communities, which demonstrate breakthrough performances in numerous high-level tasks \cite{Devlin2018,Liu2019,Sun2019,Chen2020,Dosovitskiy2020,He2021}. 
Inspired by the success of these masked DL frameworks, an intuitive and effective viewpoint, known as adversarial perturbation sparsification, is devised to design the adversarial defense framework along spectral channel in this work.
\par
\begin{figure}[t]
  \centering
   \subfigure[]{\label{fig:c1_a}\includegraphics[width=0.48\linewidth]{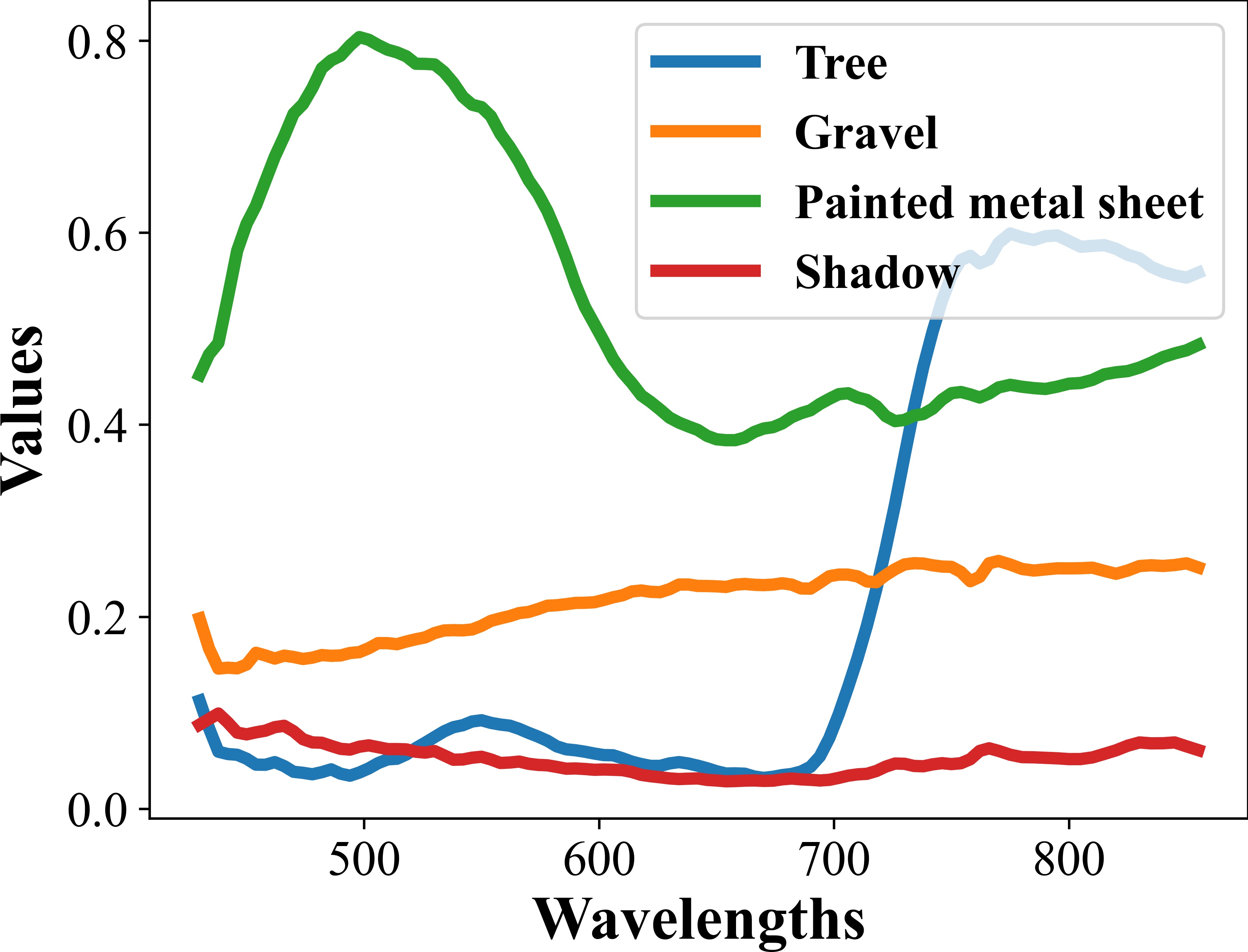}}
   \subfigure[]{\label{fig:c1_b}\includegraphics[width=0.48\linewidth]{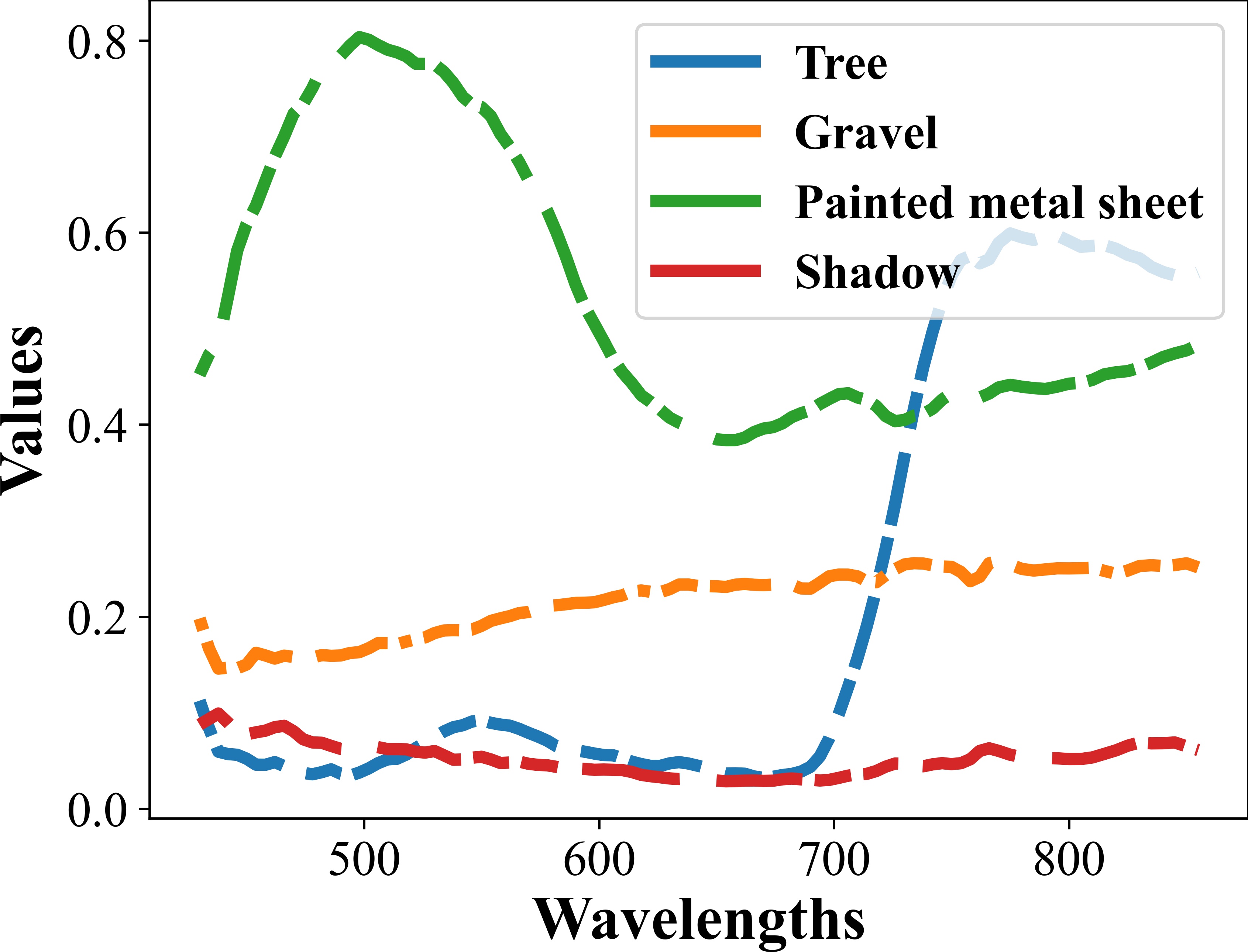}}
   \subfigure[]{\label{fig:c1_c}\includegraphics[width=0.48\linewidth]{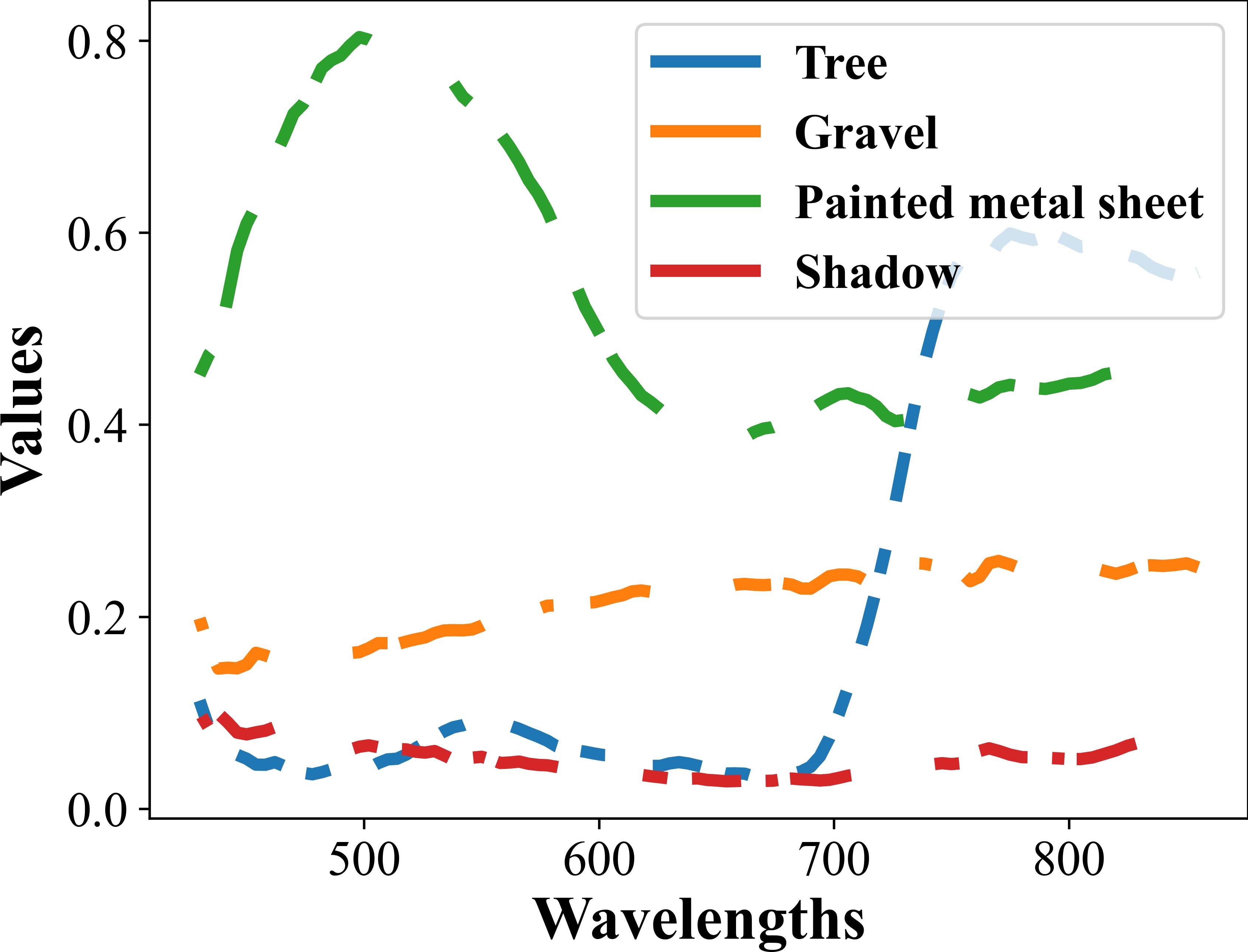}}
   \subfigure[]{\label{fig:c1_d}\includegraphics[width=0.48\linewidth]{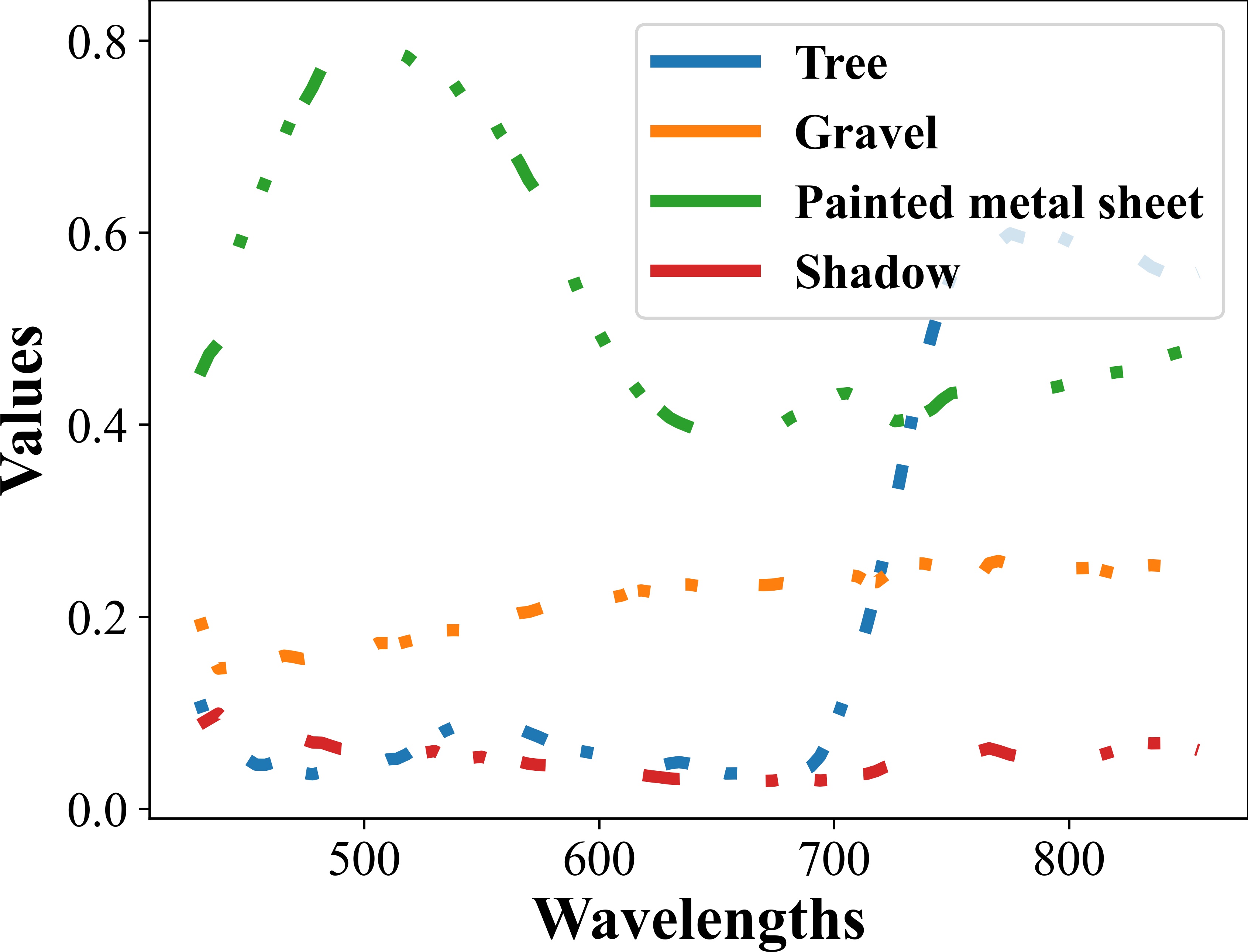}}
     \caption{Masked spectra of several land-cover objects with different masking ratios. (a) 0\%; (b) 25\%, ; (c) 50\%, ; (d) 75\%.}
  \label{fig:C1}
  \end{figure}

Adversarial perturbation sparsification tries to abandon some spectral channels of adversarial HSI by employing a random input masking strategy, for the purpose of reducing the intensity of adversarial perturbations. 
This perspective is motivated by the fact that HSIs always possess a large amount of redundant spectral information. 
As shown in Fig. \ref{fig:C1}, different ratios of input masking are conducted on the spectra of several land-cover objects.
It is still effortless to distinguish these land-cover objects according to the masked spectral signatures, even under a 75\% masking ratio. 
This phenomenon indicates that not all the spectral information is essential for hyperspectral characterization and a moderate degree of input masking might not affect HSI analysis results but achieves a promising perturbation sparsification performance. 
Furthermore, the success of hyperspectral band selection methods \cite{Yuan2017, Task2017} also verifies above ideas. 
In a sense, performing random input masking can be considered as a dynamic band selection method to refine the adversarial HSIs. 
Moreover, we can easily design a pretext task with the input masking operation, and then MSSA can be fitted in a self-supervised manner, which helps to tackle the issue of insufficient labelled samples.

\subsection{Global Pixel-wise Combination Defense}
Global spatial information was first employed to promote the performances of hyperspectral classifiers in several previous researches \cite{Xie2022,Yu2022}. 
Most recently, some work \cite{Shi2021} finds that global spatial information can also help to enhance the transferability of DL framework. 
Since robustness is verified to be associated with transferability \cite{Xu2012}, the global spatial information can be further exploited to design adversarial defense strategies in both RGB domain \cite{Qi2020} and our research. 
\par
With the help of global spatial information, we come up with a novel spatial defense principle named global pixel-wise combination defense.
It utilizes the spatial contextual information to establish the global spatial dependency for a better defensive capability. 
To be more specific, for a given pixel, the features of its neighbor pixels and itself are aggregated to yield a more robust hyperspectral characterization. 
Under this circumstance, the prediction of this pixel would be influenced by its neighbor pixels and a pixel-wise combination has formed. 
Then, the input adversarial perturbation would be dispersed among the neighbor pixels during gradient descent process, and a larger scale adversarial perturbation is required to successfully attack DL models, resulting in a better defense effect along spatial aspect.
Utilizing spatial contextual information to design defensive strategy is firstly proposed in \cite{Xu2021}, where the authors extract global contextual information from all pixels to perform long-range context encoding for enhancing the stability of HSI classifier. 
In this work, we raise a more flexible method for global contextual information extraction.
This method is on the basis of GCN architecture with the consideration of following two factors. 
On one hand, instead of treating all the pixels as neighbors, GCN is capable of extracting the long-range contextual information from a characteristic-specific neighborhood by carefully designing its adjacency matrix, which seems to be more accurate and effective. 
On the other hand, the correlations between pixels change a lot during the training process and a dynamic description of these correlations is required to further update the adjacency matrix. 
To this end, we finally come up with a distance-based graph structure learning schedule to construct a more flexible and effective GCN block, to further establish adversarial defense in spatial aspect. 

\section{Masked Spatial-Spectral Autoencoder}\label{sec:4} 
In this section, we propose a self-supervised masked spatial-spectral autoencoder for HSI and explain how it enhances the inherent robustness of HSI analysis systems. 
\begin{figure*}[t]
    \centering
     \includegraphics[width=0.99\linewidth]{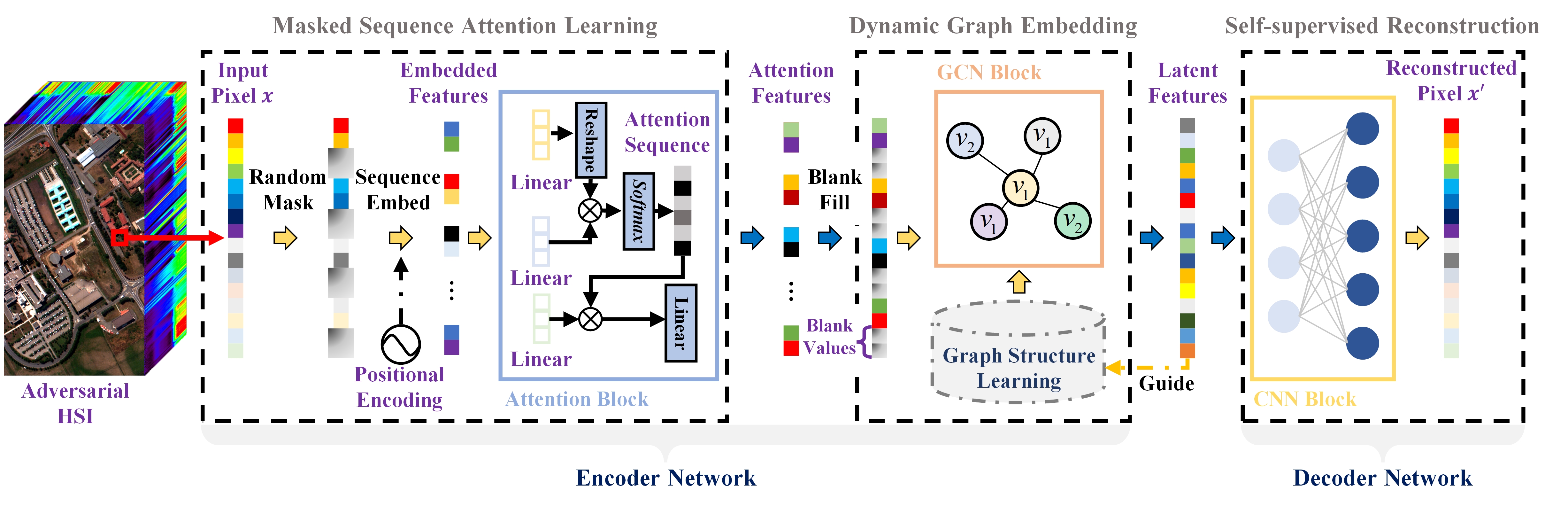}
       \caption{Flowchart of the proposed masked spatial-spectral autoencoder for hyperspectral images.}
       \label{fig:C2}
\end{figure*} 
\subsection{Overall Framework}
Based on the discussions and analyses mentioned in Section \ref{sec:3}, we propose MSSA to address the threat of adversarial attacks for HSI analysis system and its overall framework is presented in Fig. \ref{fig:C2}. 
It is effortless to find that MSSA consists of the following three modules:
\par
1) \textbf{Masked Sequence Attention Learning}. We adopt random input masking strategy to transform pixel spectra into a set of sequence patches as the inputs of self-attention block. 
In the self-attention block, the spectral correlations among these sequence patches are computed to generate the corresponding self-attention features. 
\par
2) \textbf{Dynamic Graph Embedding}. The self-attention features are filled with several trainable sequence vectors and sent to a GCN block. 
In this GCN block, the filled self-attention features among a certain global neighborhood that determined by a graph structure learning module, are aggregated to create the latent representations. 
\par
3) \textbf{Self-supervised reconstruction}. We feed the latent representations into decoder block to reconstruct masked pixel spectra for accomplishing the pretext task.
MSSA employs the reconstruction errors to design loss functions and fits the training data in a self-supervised manner. 
\par
In the following subsections, we will demonstrate the main constitutes of MSSA in detail. 
\subsection{Masked Sequence Attention Learning}
\textbf{1) Sequence random masking:} Input random masking is the core operation to perform adversarial attack from spectral aspect.
For a given pixel spectrum, it would be regarded as sequence data and divided into several regular non-overlapping patches. 
Then, a subset of patches is randomly sampled without replacement according to some certain probability distribution and the remaining ones would be masked (discarded). 
To prevent a potential region bias (i.e., the masked patches gather around several continuous bands), the uniform distribution would be selected to perform patch sampling.
\par
Mathematically, let $\bm{x}=$ $\{x_i\}_{i=1}^{c} \in \mathbb{R}^{c}$ be the sequence form of a pixel spectrum, where $c$ refers to the band number of input HSI. 
After spectrum sequence division with patch size $m$, a novel patch set $\bm{p} = \{\hat{\bm{p}}_i\}_{i=1}^{l} \in \mathbb{R}^{l \times m} $ is required, where $\hat{\bm{p}}_i = \{x_{(i-1)m},x_{(i-1)m+1},\cdots,x_{im}\}$, $l = \lceil \frac{c}{m} \rceil $ and $\lceil \cdot \rceil$ represents the ceiling operation.
In practice, zero-padding operation would be conducted on $\bm{x}$ to ensure band number $c$ could be divisible by a patch size $m$. 
With a given masking ratio $\alpha$, only $r$ spectrum patches are reserved by randomly sampling under an uniform distribution, where $ r = \lceil l(1-\alpha) \rceil$. 
Finally, the unmasked patch set $\bm{p}_{unmask} = \{\tilde{\bm{p}}_{i}\}_{i=1}^{r}$ is further sent to spectral self-attention learning (SSL) module. 
\par 
\textbf{2) Spectral self-attention learning:} Self-attention network refers to a feature extractor that can extract global contextual information and is commonly used in HSI classification and denoising tasks \cite{Shi2021, Zhang2022}. 
In this paper, a SSL mechanism is exploited for the following two purposes. 
Firstly, the self-attention mechanism can help to obtain long-range spectral context for the sake of building global spectral correlation, which further improves the intrinsic robustness in spectral aspect. 
On the other hand, since the majority of spectrum patches have been masked, only more discriminative features are extracted from the unmasked ones, could MSSA achieve a promising performance in the pretext task (masked pixel spectrum reconstruction). 
\par
As shown in Fig. \ref{fig:C2}, SSL module first converts the unmasked patches into the embedded features by a linear projection, which can be described as: 
\begin{equation} \label{eq:7}
  \bm{h}_{i}=\bm{W}_e \tilde{\bm{p}}_{i}, \quad i = 1,2,\cdots,r,
\end{equation}
where $\tilde{\bm{p}}_{i}$ denotes the $i$-th patch in unmasked patch set, $\bm{W}_{e} \in \mathbb{R}^{m \times d}$ represents the linear projection matrix figuring out $d$-dimension vectors and $\bm{h}_{i} \in \mathbb{R}^{d} $ represents the embedded features of $i$-th unmasked patch. 
It is notable that all the unmasked patches share an uniform linear projection for the purposed of reducing network parameters. 
Then, sine and cosine functions of different frequencies proposed in \cite{Devlin2018} are used to generate element-wised positional encoding vector $\bm{\phi}_e \in \mathbb{R}^{d}$ to add the position information upon above embedded features by $\hat{\bm{h}}_{i} = \bm{h}_{i}+\bm{\phi}_e$. 
In this way, SSL module is capable of distinguishing the order of embedded features. 
\par 
\begin{figure*}[t]
  \centering
   \includegraphics[width=0.99\linewidth]{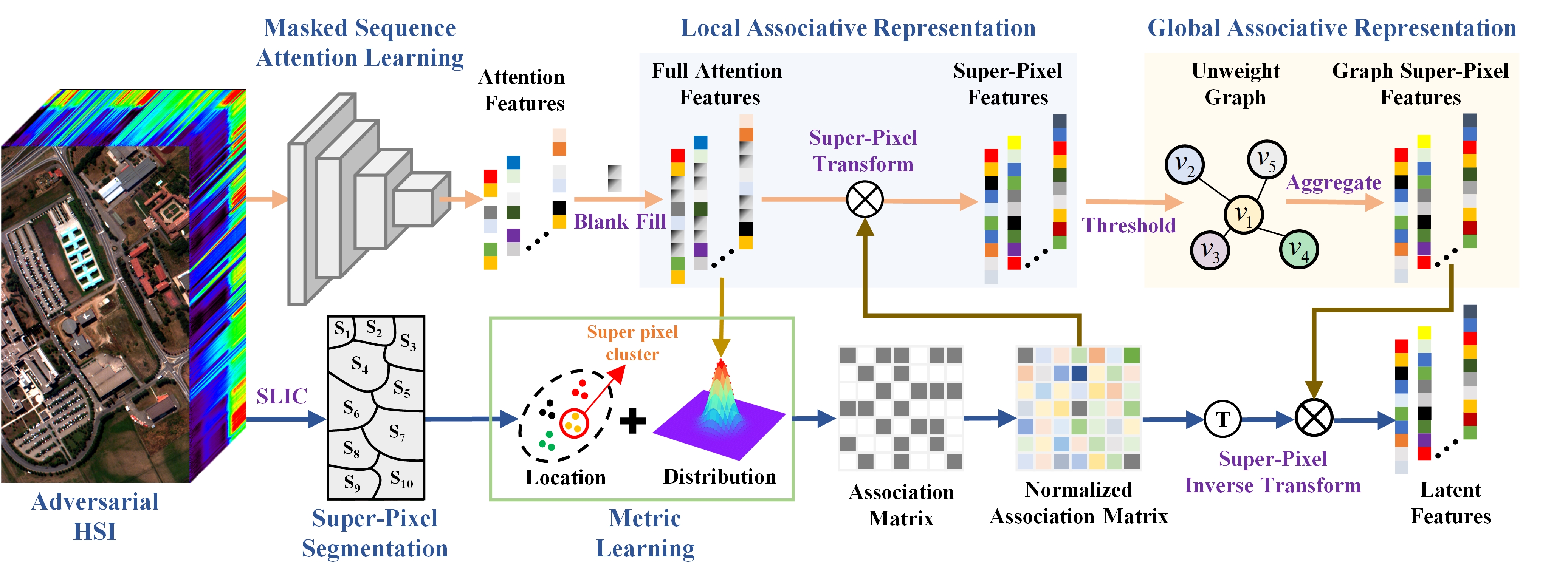}
     \caption{Illustration of the dynamic graph embedding module. Simple linear iterative clustering (SLIC) refers to a classical superpixel method.}
  \label{fig:C3}
  \end{figure*}
Subsequently, the embedded features with position information are fed into the self-attention block. 
In this block, each feature would be simultaneously multiplied with three learnable linear matrixes to generate attention component vectors.
This process can be demonstrated as follows: 
\begin{equation} \label{eq:8}
    \bm{q}_{i}=\bm{W}_{q} \hat{\bm{h}}_{i}, \
    \bm{k}_{i}=\bm{W}_{k} \hat{\bm{h}}_{i}, \
    \bm{v}_{i}=\bm{W}_{v} \hat{\bm{h}}_{i},
\end{equation}
where $\{\bm{W}_q,\bm{W}_k,\bm{W}_v\} \in \mathbb{R}^{d \times d}$ represent \{query matrix, key matrix, value matrix\}, which are shared by all the embedded features. 
$\{\bm{q}_{i},\bm{k}_{i},\bm{v}_{i}\} \in \mathbb{R}^{d}$ corresponds to \{query vector, key vector, value vector\} of the $i$-th embedded feature $\hat{\bm{h}}_{i}$, $i = 1,2,\cdots,r$. 
With Eq. (\ref{eq:8}), we can eventually transform the embedded features with position information into query matrix $\bm{Q} = \{\bm{q}_{i}\}_{i=1}^{r}$, key matrix $\bm{K} = \{\bm{k}_{i}\}_{i=1}^{r}$ and value matrix $\bm{V} = \{\bm{v}_{i}\}_{i=1}^{r}$, where $\{\bm{Q}, \bm{K}, \bm{V}\} \in \mathbb{R}^{r \times d}$. 
Furthermore, the attention feature matrix $\bm{A} = \{\bm{a}_{i}\}_{i=1}^{r} \in \mathbb{R}^{r \times d}$ can be calculated as: 
\begin{equation} \label{eq:9}
  \bm{A}=\operatorname{Attention}(\bm{Q}, \bm{K}, \bm{V})=\operatorname{softmax}\left(\frac{\bm{Q} \bm{K}^{T}}{\sqrt{d}}\right) \bm{V},
\end{equation} 
where $\operatorname{softmax}(\cdot)$ means the softmax activation function and $d$ refers to the dimension of embedded features. 
In Eq. (\ref{eq:9}), the query matrix $\bm{Q}$ and the transpose of key matrix $\bm{K}$ are first employed to compute the correlation among different embedded features with a normalization coefficient $\sqrt{d}$ and softmax function.
Then, the attention features are represented as a linear combination of all embedded features with the weight coefficients determined by the calculated correlation among different embedded features. 
Finally, in order to make these attention features more discriminative, a single linear mapping matrix $\bm{W}_{t} \in \mathbb{R}^{d \times n}$ is used to further dispose them by $\hat{\bm{A}} = \bm{A}\bm{W}_{t}$, where $\hat{\bm{A}} = \{\hat{\bm{a}}_{i}\}_{i=1}^{r} = \{\bm{a}_{i}\bm{W}_{t}\}_{i=1}^{r} \in \mathbb{R}^{r \times n}$ comes out to be the final outputs of SSL module. 
\subsection{Dynamic Graph Embedding}
To promote the robustness against adversarial attack in spatial aspect, dynamic graph embedding (DGE) devotes to building the global pixel-wise combination for a given pixel with GCN. 
\par
\textbf{1) Multi-scale associative representation:} 
Considering the unacceptable computational complexity, GCN in DGE works on superpixel-based nodes instead of pixel-based nodes. 
Enlightened by \cite{Liu2021}, we propose a superpixel-based multi-scale GCN block, as demonstrated in Fig. \ref{fig:C3}.
\par
Before implementing DGE, we first utilize several learnable blank vectors to fill the masked sequence patches.
The attention features $\hat{\bm{A}}$ and blank vectors $\{\bm{m}_i\}_{i=1}^{l-r} \in \mathbb{R}^{(l-r)\times n}$ would be concatenated with the origin patch order and reshaped into a full attention feature vector $\bm{z} \in \mathbb{R}^{t}, t = ln$. 
For each pixel $\bm{x}$ in HSI $\bm{X} \in \mathbb{R}^{h \times w \times c}$, we can compute its full attention feature vector $\bm{z}$ with SSL module and blank filling operation, ultimately generating a full attention feature set $\bm{Z} = \{\bm{z}_{i}\}_{i=1}^{hw} \in \mathbb{R}^{hw \times n}$ as the input of DGE module. 
\par
At the same time, the input HSI is divided into several spatially connected and spectrally similar superpixels by the simple linear iterative clustering (SLIC) method \cite{Achanta2012}.
Since there exists one-to-one correspondence between the full attention features and pixels in HSI, this segmentation result provides global spatial correlation for the full attention feature set. 
It is notable that we should have employed the full attention feature set to implement superpixel segmentation, instead of the input HSI. 
However, the blank filling operation with randomly initialized vectors could destroy the spatial dependency of origin attention features.
Therefore, we exploit the superpixel segmentation result derived from input HSI to approximate the one of full attention features. 
\par 
Let $\bm{S}=\left\{\bm{s}_{i}\right\}_{i=1}^{H}$ be the superpxiel set, $\bm{s}_{i} = \{\bm{z}_{j}^{i}\}_{j=1}^{N_i}$ represents the $i$-th superpixel block, $\bm{z}_{j}^{i}$ denotes the $j$-th element in $\bm{s}_{i}$, $H$ refers to the number of superpixel blocks and $N_i$ is the number of elements in $\bm{s}_{i}$, where $\bm{s}_{i} \cap \bm{s}_{j}=\emptyset, \forall i \neq j$ and $\sum_{i=1}^{H} N_{i} = hw$. 
Then, the statistical magnitudes (mean vector $\bm{\mu}_{i}$ and covariance matrix $\bm{\varGamma}_{i}$) of superpixel block $\bm{s}_{i}$ are computed as: 
\begin{equation}\label{eq:10}
\bm{\mu}_{i} = \frac{1}{N_{i}} \sum_{j=1}^{N_{i}} \bm{z}_{j}^{i},
\end{equation}
\begin{equation}\label{eq:11}
  \bm{\varGamma}_{i} =\frac{1}{N_{i}-1} \sum_{j=1}^{N_{i}}\left(\bm{z}_{j}^{i}-\bm{{\mu}}_{i}\right)\left(\bm{z}_{j}^{i}-\bm{{\mu}}_{i}\right)^{T}. 
\end{equation}
\par 
With these statistical magnitudes, a distance-based mapping mechanism is designed to achieve a local associative representation and reduce the computational complexity. 
This mechanism allows the spatial correlation to be transmitted between full attention features and superpixels. 
More precisely, an association matrix $\bm{O} \in \mathbb{R}^{hw \times H}$ between attention feature and superpixel is established as 
\begin{equation} \label{eq:12}
  \bm{O}_{i, j}=
    \begin{cases}
      \sqrt{(\bm{z}_{i}-\bm{\mu}_{j})^{T} \bm{\varGamma}_{j}^{-1}(\bm{z}_{i}-\bm{\mu}_{j})}, &\text { if } \bm{z}_{i} \in \bm{s}_{j} \\ 
      \quad \quad \quad \quad \quad \quad 0, &\text { otherwise }
    \end{cases},
\end{equation}
where $\bm{O}_{i, j}$ represents the entry of $\bm{O}$ at the location $(i, j)$.
This entry describes the distance between attention feature $\bm{z}_{i} \in \bm{s}_{j}$ and the centroid of $\bm{s}_{j}$ in a Mahalanobis space.
With Eq. (\ref{eq:12}), we could use the following formula to figure out the local associative representation $\bm{L}$ as
\begin{gather}\label{eq:13}
  \bm{L}=\tilde{\bm{O}}^{T} \bm{Z}, \quad \tilde{\bm{O}}_{i, j} = \bm{O}_{i, j} / \sum_{m} \bm{O}_{m, j},
\end{gather}
where $\bm{L} = \{\bm{l}_{i}\}_{i}^{H}$ also represents the calculated superpixel nodes, $\tilde{\bm{O}}$ denotes the normalized $\bm{O}$ along column.
Considering the nodes in $\bm{L}$ as the vertices, we can establish an undirected graph $G=(\bm{L}, \bm{E})$, where $\bm{E}$ represents the edge set among different vertices and is generally computed on the basis of radial basis function (RBF):
\begin{equation}\label{eq:14}
  \bm{E}_{i, j}(\varepsilon)=
    \begin{cases}
      1, &\text { if } \exp \left(-\frac{\left\|\bm{l}_{i}-\bm{l}_{j}\right\|^{2}}{\sigma^{2}}\right) \geq  \varepsilon\\
      0, &\text { otherwise }
    \end{cases},
\end{equation}
where $\sigma$ denotes the hyper-parameter to control the width of RBF and $\varepsilon$ represents the pre-set threshold determined as $\sum_{i}\sum_{j} {\exp \left(-\frac{\left\|\bm{l}_{i}-\bm{l}_{j}\right\|^{2}}{\sigma^{2}}\right)})/(H\times H)$ in this paper. 
Then, according to the GCN theory proposed in \cite{Liu2021}, the global associative representation $\hat{\bm{L}}$ is computed as follows:
\begin{equation}\label{eq:15}
  \hat{\bm{L}}=\Phi \left(\widetilde{\bm{D}}^{-\frac{1}{2}} \widetilde{\bm{E}} \widetilde{\bm{D}}^{-\frac{1}{2}} \bm{L} \bm{W}_{g}+\bm{b}\right),
\end{equation}
where $\tilde{\bm{E}}=\bm{E}+\bm{I}$ is denoted as the renormalization form of $\bm{E}$, $\widetilde{\bm{D}}$ refers to a diagonal matrix representing the degrees of $\tilde{\bm{E}}$ and $\widetilde{\bm{D}}_{i, i}=\sum_{j} \widetilde{\bm{A}}_{i, j}$, $\bm{W}_{g}$ is defined as the learnable weights of GCN, $\Phi (\cdot)$ means the nonlinear activation function and $\bm{b}$ represents the bias vector. 
Finally, $\hat{\bm{L}}$ is utilized to propagate the multi-scale (both local and global) spatial correlation to the full attention features with normalized association matrix $\tilde{\bm{O}}$ by 
\begin{equation}\label{eq:16}
  \hat{\bm{Z}}=\tilde{\bm{O}}\hat{\bm{L}}, 
\end{equation}
where $\hat{\bm{Z}}=\{\hat{\bm{z}}_{i}\}_{i=1}^{hw} \in \mathbb{R}^{hw \times n}$ denotes the learned embeddings of DGE module, named latent features, representing the full attention feature set with multi-scale spatial correlation.
\par
\textbf{2) Dynamic graph structure learning:} In \cite{Liu2021}, the authors find that one major disadvantage of conventional GCN refers to its fixed graph structure. 
For achieving a better graph embedding, we take advantage of the global associative representation of GCN to adjust its adjacency matrix. 
Specifically, let $\bm{E}^{\ell }$ and $\hat{\bm{L}^{\ell }}$, calculated by Eq. (\ref{eq:14}) and Eq. (\ref{eq:15}), be the adjacency matrix and global associative representation of GCN in the $\ell $-th training epoch. 
Afterward, considering the gradient backpropagation issue, we take the inner product of global associative representation to represent the learned adjacency matrix $\tilde{\bm{E}}^{\ell }$ as 
\begin{equation}\label{eq:17}
  \tilde{\bm{E}}^{\ell }=\Theta  \left(\hat{\bm{L}^{\ell }} \hat{\bm{L}^{\ell }}^{\top}\right),
\end{equation}
where $\Theta(x) = 1 /\left(1+e^{-x}\right)$ represents the sigmoid function to perform value normalization. 
Then, $\tilde{\bm{E}}^{\ell }$ is utilized to update the original adjacency matrix $\bm{E}^{\ell }$ in a linear combination form, with a hyper-parameter $\beta $ mediating its impact as:  
\begin{equation}\label{eq:18}
  \hat{\bm{E}^{\ell }}=\beta \tilde{\bm{E}}^{\ell }+(1-\beta) \bm{E}^{\ell },
\end{equation}
where $\hat{\bm{E}^{\ell }}$, a weighted adjacency matrix, represents the updating result.
In a similar manner, $\hat{\bm{E}^{\ell }}$ should also be transformed into an unweighted adjacency matrix by a certain threshold $\gamma $, and then we can get the adjacency matrix of GCN in the $(\ell+1)$-th training epoch as:
\begin{equation}\label{eq:19}
  \bm{E}^{\ell+1 }_{i,j} = \hat{\bm{E}^{\ell }}_{{i,j}}(\gamma) = 
  \begin{cases}
    1, &\text { if } \hat{\bm{E}^{\ell }}_{{i,j}} \geq  \gamma\\
    0, &\text { otherwise }
  \end{cases}.
\end{equation}
\par 
Furthermore, to restrict the structure similarity between $\bm{E}^{\ell+1 }$ and $\bm{E}^{\ell }$, the structure sparsity is used as a regular term via $L_{1}$ norms. 
In this way, the threshold $\gamma $ can be determined by solving the following optimization problem:
\begin{equation}\label{eq:20}
  {\gamma}^{\star} =\arg \min_{\gamma \in \{0,1\}} ({\|\hat{\bm{E}^{\ell }}(\gamma)\|_{1}-\|\bm{E}^{\ell }\|_{1}}),
\end{equation}
where $\|\cdot\|_{1}$ is the $l_1$-norm of a matrix. 

\begin{algorithm}[!t]
  \caption{Masked Spatial-Spectral Autoencoder}\label{algorithm:01}
  \LinesNumbered
  \KwIn {HSI $\bm{X}$, Training epoch $E$, Superpixel segmentation method SLIC}
  \KwOut {Multi-scale associative representation $\hat{\bm{Z}}$}
  \tcp{\footnotesize Sequence random masking}
  Generate the unmasked patch set $\bm{P}_{unmask}$ for HSI $\bm{X}$ with sequence random masking operation. \\
  Calculate the superpixel set $\bm{S}$ for HSI $\bm{X}$ with SLIC method. \\
  \For{$e=0$ to $E$}{
    \tcp{\footnotesize Spectral self-attention learning}
  Compute the spectral self-attention features $\hat{\bm{A}}$ using Eq. (\ref{eq:7}) to Eq. (\ref{eq:9}) based on $\bm{P}_{unmask}$. \\
  Calculate the local associative representation $\bm{L}$ using Eq. (\ref{eq:10}) to Eq. (\ref{eq:13}) according to $\bm{S}$. \\
      \tcp{\footnotesize Dynamic graph structure learning}
      \eIf{$e = 0$}
      {
      Generate the undirected graph $G = (\bm{L}, \bm{E}_e)$ with Eq. (\ref{eq:14}) based on $\bm{L}$.
      }
      {
      Calculate the learned adjacency matrix $\tilde{\bm{E}}_{e-1}$ with Eq. (\ref{eq:17}) based on $\hat{\bm{L}}_{e-1}$.\\
      Update the undirected graph $G = (\bm{L}, \bm{E}_e)$ using Eq. (\ref{eq:18}) to Eq. (\ref{eq:20}) by $\bm{E}_{e-1}$ and $\tilde{\bm{E}}_{e-1}$.  
      }
      Compute the global associative representation $\hat{\bm{L}}_{e}$ by Eq. (\ref{eq:15}) with $G$ and $\bm{L}$.\\
  \tcp{\footnotesize Multi-scale associative representation}
  Generate the multi-scale associative representation $\hat{\bm{Z}}$ using Eq. (\ref{eq:16}) according to $\hat{\bm{L}}_{e}$.\\
  \tcp{\footnotesize Masked sequences reconstruction}
  Calculate the reconstructed HSI $\hat{\bm{X}}$ with Eq. (\ref{eq:21}) to Eq. (\ref{eq:22}) based on $\hat{\bm{Z}}$.\\
  Compute the reconstruction loss $\mathcal{L}$ between $\hat{\bm{X}}$ and $\bm{X}$.\\
  \tcp{\footnotesize Self-supervised optimization}
  Do one training step with reconstruction loss $\mathcal{L}$.
  }
  {\bf return} $\hat{\bm{Z}}$
\end{algorithm}

\subsection{Self-supervised reconstruction}
\textbf{1) Masked sequences reconstruction:} Figuring out the masked sequence based on the generated latent features $\hat{\bm{Z}}$ is the last mission to accomplish the pretext task. 
Different from BERT \cite{Devlin2018}, whose decoder is simply a multi-layer perceptron (MLP), 1-D convolutional neural network (CNN) is applied to design a nontrivial decoder for MSSA, since reconstructing pixel spectrum has a lower semantic level than predicting missing words \cite{He2021}.
To this end, a stack of cascaded 1-D CNN layers are employed to establish a CNN block and the output of $i$-th layer in this block can be defined as
\begin{equation}\label{eq:21}
  \bm{Y}^{i}=
    \begin{cases}
      f\left(\bm{W}^{i}_{c} * \hat{\bm{Z}}+\bm{b}^{i}_{c}\right), & i=1 \\ 
      f\left(\bm{W}^{i}_{c} * \bm{y}^{(t-1)}+\bm{b}^{i}_{c}\right), & i \geq 2
    \end{cases},
\end{equation}
where $f$ denotes the nonlinear activation function, $*$ represents the convolution math operation, $\bm{W}^{i}_{c}$ and $\bm{b}^{i}_{c}$ are the weight matrix and bias term of the $i$-th layer, $i = 1,2,\cdots, C$.
Finally, a fully connected layer transforms the hidden feature $\bm{Y}^{C}$ refined by the CNN block into reconstruction result $\hat{\bm{X}}$: 
\begin{equation} \label{eq:22}
  \hat{\bm{X}} = \Theta  \left(\bm{W}_{r}\bm{Y}^{C}+\bm{b}_{r}\right),
\end{equation}
where $\Theta(\cdot)$ means the sigmoid function, $\bm{W}_{r}$ and $\bm{b}_{r}$ represent the parameters of fully connected layer.
Significantly, sigmoid function $\Theta(\cdot)$ in Eq. (\ref{eq:22}) is used to restrict the reconstruction result among a more appropriate value range, which contributes a lot to the speediness of convergence.
\par
\begin{table*}[!t]
  \footnotesize
    \centering
    \caption{{The detailed information of training/testing samples for experiments.}}
    \begin{tabular}{|c | c | c| c | c| c| c| c |c| c|}
    \hline
    \multirow{2}{*}{\textbf{ID}}  & \multicolumn{3}{c|}{\bf Pavia University} & \multicolumn{3}{c|}{\bf Salinas Scene}  & \multicolumn{3}{c|}{\bf Houston2013}\\
  \cline{2-10}
    & Class Name & Train & Test & Class Name & Train & Test & Class Name & Train & Test \\
    \hline\hline
    C1 & Asphalt &200 & 6431 & Brocoli-green-weeds-1 & 200 & 1809 & Grass-healthy & 200 & 1051 \\
    C2 & Meadows&200 & 18449 & Brocoli-green-weeds-2&200 & 3526& Grass-stressed& 200 & 1054\\
    C3 &Gravel& 200 & 1899 & Fallow&200 & 1776& Grass-synthetic & 200 & 497\\
    C4 & Trees & 200 & 2864& Fallow-rough-plow&200 & 1194& Tree& 200 & 1044\\
    C5 &Metal sheet& 200 & 1145&Fallow-smooth& 200 & 2478&Soil&  200 & 1042\\
    C6 &Bare soil& 200 & 4829& Stubble&200 & 3759&Water&  200 & 125\\
    C7 &Bitumen& 200 & 1130& Celery&200 & 3379& Residential & 200 & 1068\\
    C8 &Brick& 200 & 3482& Grapes-untrained&200 & 11071& Commercial&  200 & 1044\\
    C9 &Shadow& 200 & 747& Soil-vinyard-develop&200 & 6003&Road&  200 & 1052\\
    C10 && & &Corn-senesced-green-weeds&200 & 3078&Highway& 200 & 1027\\
    C11 && & &Lettuce-romaine-4wk&200 & 868 &Railway & 200 & 1035\\
    C12 && & &Lettuce-romaine-5wk&200 & 1727 &Parking-lot1& 200 & 1033\\
    C13 && & &Lettuce-romaine-6wk &200 & 716 &Parking-lot2& 200 & 269\\
    C14 && & &Lettuce-romaine-7wk & 200& 870 &Tennis-court& 200 & 228\\
    C15 && & &Vinyard-untrained&200 & 7068 &Running-track& 200 & 460\\
    C16 && & &Vinyard-vertical-trellis&200 & 1607 &&  &  \\
  \hline
  Total & & 1800 & 40976 &  & 3200 & 50929 & & 3000 & 12029\\
    \hline
    \end{tabular}
    \label{table1:dataset}
  \end{table*}
\textbf{2) Self-supervised optimization:} The pretext task for MSSA is to reconstruct the input HSI by predicting spectrum values for the masked sequences. 
Consequently, the reconstruction errors would be exploited as the loss function to fit MSSA model without any label information. 
So as to promoting the transferability of MSSA while reducing the computational consumption, we merely compute the reconstruction error on masked spectrum sequences.
In this way, MSSA will be trained in a self-supervised manner.
For a more precise reconstruction error, mean squared error (MSE) and spectral angle (SA) distance, describing the distinction between input HSI and reconstructed HSI in the numerical and shape aspects respectively, are used to represent the loss function as
\begin{equation}\label{loss}
  \begin{split}
    \mathcal{L} &= \mathcal{L}_{MSE}(\hat{\bm{X}}^{mask},\bm{X}^{mask}) + \zeta \mathcal{L}_{SA}(\hat{\bm{X}}^{mask},\bm{X}^{mask}) \\
                &= \sum_{i=1}^{h} \sum_{j=1}^{w} (\|\hat{\bm{X}}_{i,j}^{mask}-\bm{X}_{i,j}^{mask}\| \\
                & \quad \quad \quad \quad \quad \quad \quad + \frac{\zeta}{\pi} \arccos \frac{\hat{\bm{X}}_{i,j}^{mask} \odot  \bm{X}_{i,j}^{mask}}{\left\|\hat{\bm{X}}_{i,j}^{mask}\right\|_{2}\left\|\bm{X}_{i,j}^{mask}\right\|_{2}})
  \end{split},
\end{equation}
where $\hat{\bm{X}}^{mask}$ and ${\bm{X}}^{mask}$ are the masked part of input HSI and reconstructed HSI, $\zeta$ represents the weight coefficient of $\mathcal{L}_{SA}$ in the overall loss function, $\odot$ denotes the inner product operation of vectors. 
Once the pre-training process is complete, only the encoder network of MSSA is transferred to downstream tasks, for producing discriminative and robust spectrum representations.

\section{experiment}\label{sec:5}
In this section, intensive experiments are performed to demonstrate the effectiveness of proposed method. 
Land-cover classification is specified as the downstream task.
At the very beginning, we introduce the datasets employed in this work.
Then, the experimental setups, results and corresponding analyses are detailed in the remainder of this section.
\subsection{Datasets}
To further evaluate the performance of proposed method, we conduct experiments over three hyperspectral benchmark datasets.
\par
The first dataset named Pavia University (PaviaU) \cite{Data1}, consisting of $610\times 340$ pixels, was acquired by the reflective optics system imaging spectrometer (ROSIS-3) sensor in Pavia, Northern Italy. 
42,776 samples are labelled and divided into 9 land-cover categories while 103 spectral bands ranging from 0.43 to 0.86 $\mu m$ are selected for experiments.
The second dataset is Salinas Scene \cite{Data1}, which is collected by the 224-band airborne visible infrared imaging spectrometer (AVIRIS) sensor over Salinas Valley, California. 
The spectral range of this dataset covers from 0.4 to 2.5 $\mu m$ and 54,129 samples of 16 classes are labelled for experiments.
The last dataset refers to Houston2013 \cite{Data2} that contains 144 spectral bands and 15,011 labelled samples among 16 land-cover categories. 
This dataset, covering 0.38 to 1.05 $\mu m$, is published in 2013. 
Table \ref{table1:dataset} lists the detailed information of training/testing samples.

\subsection{Experimental Designs and Setups}
All the experiments can be divided into three parts.
The first part evaluates the robustness of different hyperspectral classifiers. 
The second part is employed to prove the effectiveness of MSSA in comparison with several representative adversarial defense strategies.
Finally, we conduct a comprehensive ablation study to investigate how each component of MSSA affects the adversarial defense performances.
\par
To better assess the resistance of DL methods, three adversarial attack methods (FGSM \cite{Goodfellow2014}, BIM \cite{Kurakin2016} and DeepFool \cite{Dezfooli2015}) are employed as the attackers. 
Noticeably, all the adversarial attack methods are modified as untargeted attack methods to achieve a better attack effect and their parameter settings are listed in Table \ref{table:2}. 
Besides, to avoid other DL frameworks from affecting the adversarial defense performance of MSSA, a naive fully connected network that also denotes as the baseline, is employed to be the classifier of MSSA. 
As for the evaluation metrics, overall accuracy (OA), average accuracy (AA), and kappa coefficient ($\kappa$) are utilized to provide quantitative analyses of comparison results.
In this work, we repeat each experiment 25 times by randomly sampling 200 training samples from the employed datasets, while both the means and standard deviations of experimental results are demonstrated. 
In terms of hyper-parameters, the balance parameter $\beta $ in dynamic graph structure learning and the weight coefficient $\zeta$ in Eq. (\ref{loss}) are empirically set as 0.1 and 0.05. 
Meanwhile, the number of superpixels for SLIC \cite{Achanta2012} method is set as 1\% of the total pixel number for all the datasets. 
Furthermore, the masking ratio $\alpha$ is set as 0.5 to make a trade-off between efficiency and effectiveness.
\par 
The Adam \cite{Kingma2014} algorithm is employed as the optimizer with an excellent learning rate scheduler \cite{He2018} and a weight decay rate of $5\times10^{-5}$.
Furthermore, we set the training epoch as 500 for MSSA. 
For a fair comparison, all the compared method are performed by using the hyper-parameters mentioned in the original paper.
Finally, the related experiments are performed with the support of an Intel(R) Core (TM) i9-10920X CPU machine and two NVIDIA GeForce RTX 3090 GPUs. 
\subsection{Performances of Hyperspectral Classifiers}
In this subsection, we evaluates the robustness of proposed method with several state-of-the-art hyperspectral classifiers, including CNN-based methods, i.e., 1-D CNN \cite{Hu2015}, 3-D CNN \cite{Hamida2018} and Hybrid CNN (HybridSN) \cite{Roy2020}; Sequence-based method, i.e., Deep RNN (DRNN) \cite{Mou2017}, SpectralFormer \cite{Hong2022}; Graph-based method, i.e., WFCG \cite{Dong2022}. 
Besides, a self-attention context network (SACNet) \cite{Xu2021} using global contextual information to confront adversarial attacks is also employed as a compared method. 
Furthermore, the baseline in this paper is a naive classifier consists of four fully connected layers.
\par
Table \ref{table:3} to Table \ref{table:5} provide the quantitative comparison results about classification performances. 
It can be noted that the baseline, which consists of plain DL frameworks and fails to generate discriminative features, achieves the poorest classification results under adversarial attacks. 
This demonstrates that a more powerful feature extractor is required to enhance the robustness of DL models. 
However, simply combining spectral and spatial information usually cannot provide expected performance.
For example, the best OA value of 3-D CNN and HybridSN is no more than 17\%.
1-D CNN and DRNN, employing DL structures to refine the information along spectral channel, have mildly alleviated the impact of adversarial attacks. 
Meanwhile, WFCG achieves an analogical effect by extracting local spatial context information. 
These three methods outperform 3-D CNN and HybridSN more than 35\% in the OA metric under FGSM attack. 
Nevertheless, when attack methods are more powerful (such as DeepFool), the local spatial or spectral contextual information is not sufficient to confront the adversarial attack.
Then, 1-D CNN, DRNN and WFCG would also be fooled by these methods. 
Based on these results we can conclude that, the majority of state-of-the-art DL classifiers are vulnerable to adversarial attacks. 
Different from aforementioned classifiers, SACNet designs a self-attention learning based global context encoding mechanism to achieve the competitive classification performances under all the attack methods over three benchmark datasets. 
In Houston2013 dataset under DeepFool attack, it even yields an OA of about 58\%, while 1-D CNN, DRNN and WFCG can only get 12.05\%, 13.02\% and 16.22\%. 
This phenomenon indicates that global spatial contextual information can contribute to address the threat of adversarial attacks in classification tasks, since making full use of global spatial contextual information refers to the main defense strategy of SACNet. 
But compared to the performances on clean datasets, there remain distinct performance degradations of SACNet after being attacked. 
\par
\begin{table}[!t]
  \renewcommand{\arraystretch}{1.5}
  \footnotesize
  \caption{The parameter settings of employed adversarial attack methods.$\epsilon$ and $\alpha$ are maximum perturbation and step size.}\label{table:2}
  \begin{tabular}{cccccc}
    \rowcolor{black!25}Attack Method & Parameters & Perturbation Norm & Target Class \\
    \rowcolor{black!10}FGSM &$\epsilon=16/255$  & $\ell_{2 } $ & None \\
    \rowcolor{black!15} &$\epsilon=16/255$  &  &  \\
    \rowcolor{black!15}&$\alpha=8/255$  &  &  \\
    \rowcolor{black!15}\multirow{-3}*{BIM}&Epoch=20  &\multirow{-3}*{$\ell_{\infty } $}  &\multirow{-3}*{None}  \\
    \rowcolor{black!5} &Overshoot=0.02 &  &  \\
    \rowcolor{black!5} \multirow{-2}*{DeepFool}&Epoch=40 &\multirow{-2}*{$\ell_{\infty } $}  &\multirow{-2}*{None}  \\
  \end{tabular}
\end{table}
\begin{table*}[t]
  \renewcommand{\arraystretch}{1.25}
  \scriptsize
  \begin{center}
  \caption{Quantitative classification results of PaviaU dataset. The \textbf{bold entries} represent the best performance in each row. }
\label{table:3}
\begin{tabular}{| c | c | c | c | c | c | c | c | c | c | c |}
\hline
{Method}     &   {Metrics}     &  {Baseline} &  {1-D CNN}  & {3-D CNN} & {HybridSN} & {DRNN} & {SpectralFormer} & {WFCG} & {SACNet} & {\cellcolor{lightgray!50} MSSA}\\
\hline\hline
\multirow{3}{*}{Clean}& {OA(\%)}   & 84.42±1.56 & 79.14±4.1 & 97.85±1.79 & 98.75±1.17 & 98.83±1.15 & 87.94±2.83 & \textbf{99.63±0.31} & 96.17±2.85 & 98.71±0.91 \\
& {AA(\%)}   & 84.64±3.48 & 82.79±3.91 & 89.95±2.15 & 95.82±3.07 & 95.25±1.47 & 89.34±1.97 & \textbf{99.49±0.39} & 88.81±4.83 & 97.79±1.4 \\
& {$\kappa$ (\%)}    & 82.95±2.93 & 76.82±3.5 & 93.93±1.8 & 96.45±2.46 & 96.72±3.3 & 83.89±1.96 & \textbf{99.51±0.04} & 88.67±1.71 & 97.53±1.53 \\
 \hline\hline
\multirow{3}{*}{FGSM}& {OA(\%)}   & 3.33±1.16 & 46.28±4.65 & 14.13±4.88 & 8.91±3.87 & 34.04±2.6 & 40.85±4.05 & 39.04±1.89 & 57.56±2.02 & \textbf{81.87±4.85} \\
& {AA(\%)}   & 11.49±2.46 & 38.96±1.92 & 29.28±1.58 & 21.82±4.15 & 42.07±2.99 & 49.91±1.14 & 57.63±3.64 & 45.46±1.68 & \textbf{79.04±5.91 }\\
& {$\kappa$ (\%)}  & 1.29±1.12 & 29.35±4.55 & 9.07±2.53 & 4.69±4.53 & 20.81±1.84 & 27.32±1.69 & 29.83±2.45 & 46.27±4.63 & \textbf{75.13±1.93} \\
 \hline\hline
\multirow{3}{*}{BIM}& {OA(\%)}    & 2.16±1.48 & 44.52±2.41 & 13.94±4.19 & 9.07±4.93 & 29.01±2.29 & 17.94±1.66 & 20.77±4.86 & 57.44±4.02 & \textbf{76.93±1.13} \\
& {AA(\%)}   & 6.98±1.88 & 37.91±4.24 & 29.07±2.77 & 22.42±3.88 & 38.93±2.46 & 34.26±1.02 & 38.51±1.14 & 45.49±1.53 & \textbf{69.16±4.62} \\
& {$\kappa$ (\%)}   & 1.05±0.56 & 27.22±2.85 & 7.12±2.17 & 4.73±2.74 & 15.67±1.21 & 13.89±4.78 & 8.85±3.92 & 46.07±2.51 & \textbf{68.53±1.81} \\
 \hline\hline

 \multirow{3}{*}{DeepFool}& {OA(\%)}  &  1.08±0.46 & 25.03±1.18 & 11.12±3.03 & 7.67±1.2 & 17.85±4.1 & 12.28±1.77 & 9.95±1.82 & 56.89±1.11 & \textbf{73.11±3.04} \\
 & {AA(\%)}   & 3.48±1.38 & 29.57±4.97 & 26.36±2.99 & 19.35±3.65 & 32.31±3.47 & 29.82±3.38 & 27.8±4.08 & 44.58±3.89 & \textbf{65.18±2.67} \\
 & {$\kappa$ (\%)}  &  0.96±0.04 & 6.73±1.17 & 2.82±2.66 & 2.91±1.91 & 4.79±2.54 & 11.75±4.16 & 6.06±3.37 & 45.27±2.41 & \textbf{63.2±1.06} \\
\hline
\end{tabular}
\end{center}
\end{table*}

\begin{table*}[t]
  \renewcommand{\arraystretch}{1.25}
  \scriptsize
  \begin{center}
  \caption{Quantitative classification results of Salinas dataset. The \textbf{bold entries} represent the best performance in each row.}
\label{table:4}
\begin{tabular}{| c | c | c | c | c | c | c | c | c | c | c |}
\hline
{Method}     &   {Metrics}     &  {Baseline} &  {1-D CNN}  & {3-D CNN} & {HybridSN} & {DRNN} & {SpectralFormer} & {WFCG} & {SACNet} & {\cellcolor{lightgray!50} MSSA}\\
\hline\hline
\multirow{3}{*}{Clean}& {OA(\%)}   & 84.79±5.49 & 82.78±5.32 & 95.22±3.44 & 96.53±3.01 & 94.26±5.9 & 91.41±4.43 & \textbf{99.52±0.37} & 90.74±1.57 & 97.78±1.63 \\
& {AA(\%)}    & 85.23±4.37 & 84.38±6.31 & 93.2±5.22 & 95.18±1.21 & 96.37±3.24 & 95.71±1.58 & \textbf{99.55±0.18} & 82.11±2.78 & 98.03±1.87 \\
& {$\kappa$ (\%)}    & 83.57±1.86 & 78.25±2.94 & 93.26±4.24 & 95.09±6.61 & 93.62±6.09 & 90.43±6.02 & \textbf{99.46±0.16} & 86.54±4.72 & 96.59±2.77 \\
 \hline\hline
\multirow{3}{*}{FGSM}& {OA(\%)}    & 15.91±4.09 & 47.94±5.59 & 13.36±6.8 & 12.82±3.8 & 40.83±5.01 & 44.12±3.97 & 46.53±4.99 & 52.99±4.09 & \textbf{83.21±1.54} \\
& {AA(\%)}    &22.87±1.95 & 47.72±2.2 & 18.39±1.8 & 18.99±2.88 & 51.31±4.7 & 52.02±5.09 & 63.75±2.87 & 49.16±3.02 & \textbf{83.86±5.39} \\
& {$\kappa$ (\%)}  & 3.78±1.99 & 41.93±2.87 & 4.86±2.88 & 4.27±4.04 & 34.84±2.88 & 38.04±4.74 & 41.32±6.14 & 48.23±3.64 &\textbf{81.27±4.43}  \\
 \hline\hline
\multirow{3}{*}{BIM}& {OA(\%)}    & 11.11±2.13 & 46.79±2.52 & 12.98±5.81 & 13.16±3.19 & 40.14±3.91 & 32.13±3.51 & 39.66±6.92 & 53.15±2.9 & \textbf{78.93±4.81} \\
& {AA(\%)}   & 14.31±1.37 & 46.54±4.29 & 18.02±2.59 & 19.19±5.39 & 50.95±5.81 & 39.86±5.41 & 45.15±5.86 & 49.13±2.92 & \textbf{81.57±2.11} \\
& {$\kappa$ (\%)}    & 1.93±1.03 & 40.71±2.11 & 4.28±3.59 & 4.53±2.39 & 34.09±2.26 & 24.97±3.27 & 34.17±2.52 & 48.4±5.88 & \textbf{76.26±1.18 } \\
 \hline\hline
 \multirow{3}{*}{DeepFool}& {OA(\%)}    & 7.99±2.76 & 29.52±6.86 & 5.21±2.61 & 4.96±5.13 & 22.28±6.21 & 15.73±1.75 & 13.91±4.64 & 48.29±2.01 & \textbf{72.17±1.93} \\
 & {AA(\%)}   &  8.07±6.17 & 34.69±4.21 & 8.98±6.05 & 8.43±6.33 & 32.51±3.34 & 21.11±4.73 & 15.72±4.97 & 47.13±1.74 & \textbf{67.73±6.49} \\
 & {$\kappa$ (\%)}   & 0.09±0.02 & 22.18±2.97 & 0.44±0.14 & 0.28±0.08 & 14.76±6.51 & 7.37±3.76 & 5.68±2.79 & 43.11±5.71 & \textbf{68.56±3.15} \\
\hline
\end{tabular}
\end{center}
\end{table*}

\begin{table*}[t]
  \renewcommand{\arraystretch}{1.25}
  \scriptsize
  \begin{center}
  \caption{Quantitative classification results of Houston2013 dataset. The \textbf{bold entries} represent the best performance in each row.}
\label{table:5}
\begin{tabular}{| c | c | c | c | c | c | c | c | c | c | c |}
\hline
{Method}     &   {Metrics}     &  {Baseline} &  {1-D CNN}  & {3-D CNN} & {HybridSN} & {DRNN} & {SpectralFormer} & {WFCG} & {SACNet} & {\cellcolor{lightgray!50} MSSA}\\
\hline\hline
\multirow{3}{*}{Clean}& {OA(\%)}   & 74.79±3.42 & 81.53±3.99 & 99.73±3.23 & 99.91±0.04 & \textbf{99.92±0.07} & 86.41±3.56 & 92.43±3.4 & 99.71±0.08 & 97.92±1.48 \\
& {AA(\%)}    & 75.23±3.86 & 81.39±3.95 & 88.93±2.1 & 96.67±2.09 & \textbf{96.97±3.05} & 87.81±2.85 & 92.17±3.64 & 92.17±3.37 & 95.41±2.01 \\
& {$\kappa$ (\%)}    & 73.56±3.75 & 78.97±2.5 & 94.06±2.21 & 97.93±1.45 & \textbf{98.29±0.34} & 85.27±2.02 & 91.81±2.6 & 93.44±2.77 & 94.57±3.95 \\
 \hline\hline
\multirow{3}{*}{FGSM}& {OA(\%)}    & 14.02±3.47 & 30.02±2.79 & 11.73±3.02 & 15.75±2.86 & 35.09±2.99 & 32.76±2.87 & 45.07±2.04 & 59.29±3.86 & \textbf{79.51±2.93} \\ 
&AA(\%) & 15.96±3.02 & 30.67±3.61 & 9.37±3.19 & 13.05±3.01 & 32.24±3.69 & 36.04±3.77 & 47.59±3.25 & 51.59±2.99 & \textbf{80.18±2.53} \\
&Kappa(\%) & 7.3±3.96 & 24.3±3.48 & 4.3±2.47 & 8.4±3.87 & 29.6±3.56 & 28.45±2.75 & 40.47±3.98 & 56.19±3.31 & \textbf{77.81±2.84} \\ 
 \hline\hline
\multirow{3}{*}{BIM}& {OA(\%)}   & 9.86±3.04 & 27.84±3.85 & 11.64±2.3 & 16.58±3.62 & 27.08±3.36 & 12.18±2.79 & 29.34±2.05 & 59.9±2.36 & \textbf{71.36±2.93} \\ 
&AA(\%) & 8.28±1.45 & 28.4±3.92 & 9.32±2.02 & 13.63±3.49 & 23.89±3.24 & 10.77±3.57 & 29.99±2.71 & 52.04±3.87 & \textbf{72.86±2.71} \\
&Kappa(\%) & 3.00±0.99 & 22.02±3.46 & 4.1±3.51 & 9.32±2.56 & 20.8±3.96 & 6.98±3.61 & 23.29±2.95 & 56.74±3.2 & \textbf{69.05±3.87} \\
 \hline\hline

 \multirow{3}{*}{DeepFool}& {OA(\%)}   & 3.12±3.65 & 12.05±2.03 & 4.39±2.45 & 8.29±3.22 & 13.02±3.6 & 4.93±3.56 & 16.22±3.67 & 58.88±3.25 & \textbf{70.47±3.76} \\ 
 &AA(\%) & 3.14±2.54 & 13.41±3.03 & 3.44±2.95 & 6.71±2.88 & 10.77±3.89 & 4.23±2.89 & 15.32±3.83 & 51.01±2.04 & \textbf{72.03±3.37} \\ 
 &Kappa(\%) & 0.42±0.19 & 5.1±2.77 & 1.6±0.64 & 3.17±2.13 & 5.43±3.61 & 2.67±2.27 & 8.91±3.12 & 55.61±2.67 & \textbf{68.03±3.07} \\
\hline
\end{tabular}
\end{center}
\end{table*}
\begin{figure*}[!t]
  \centering
   \subfigure[]{\label{fig:d1_a}\includegraphics[width=0.16\linewidth]{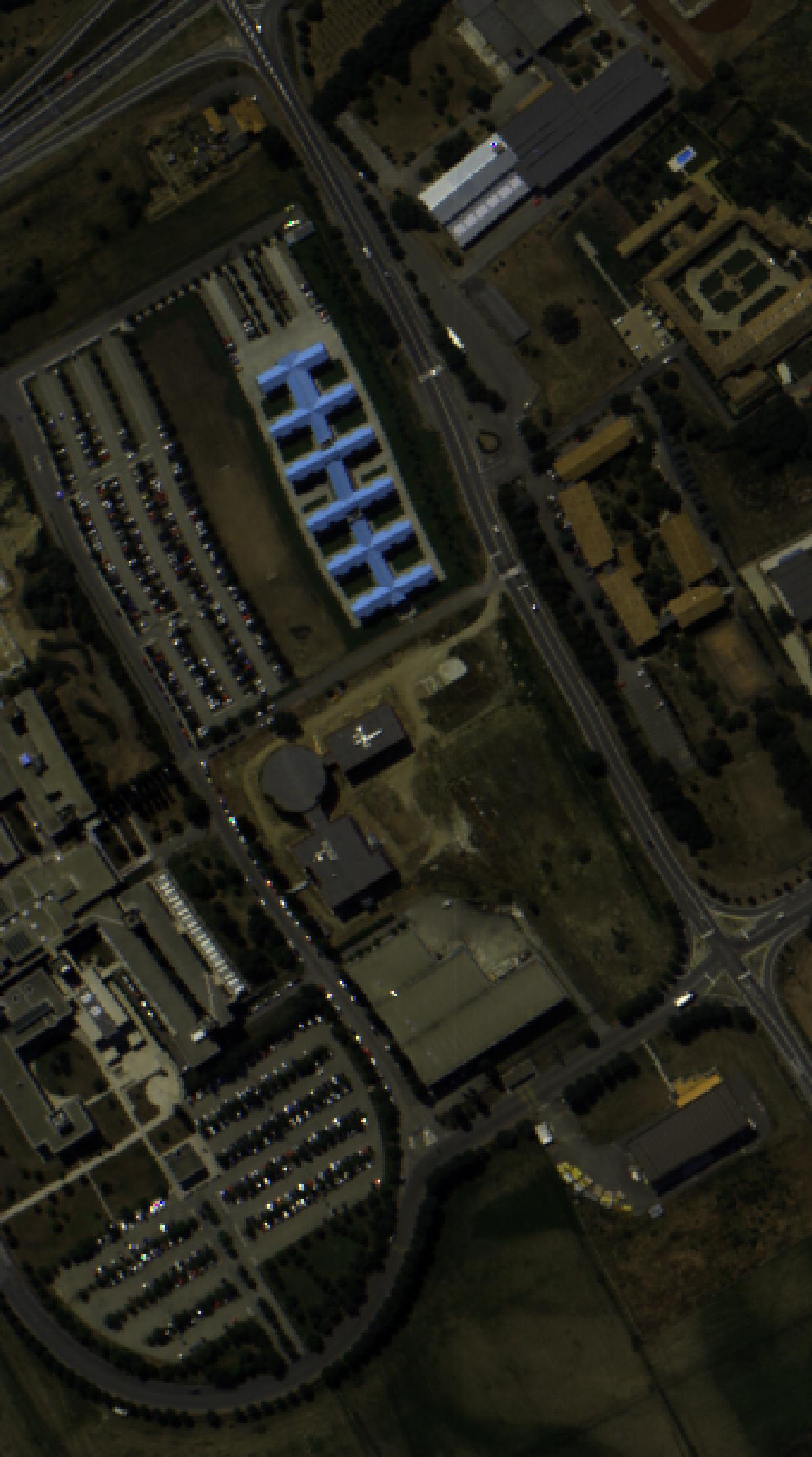}}
   \subfigure[]{\label{fig:d1_b}\includegraphics[width=0.16\linewidth]{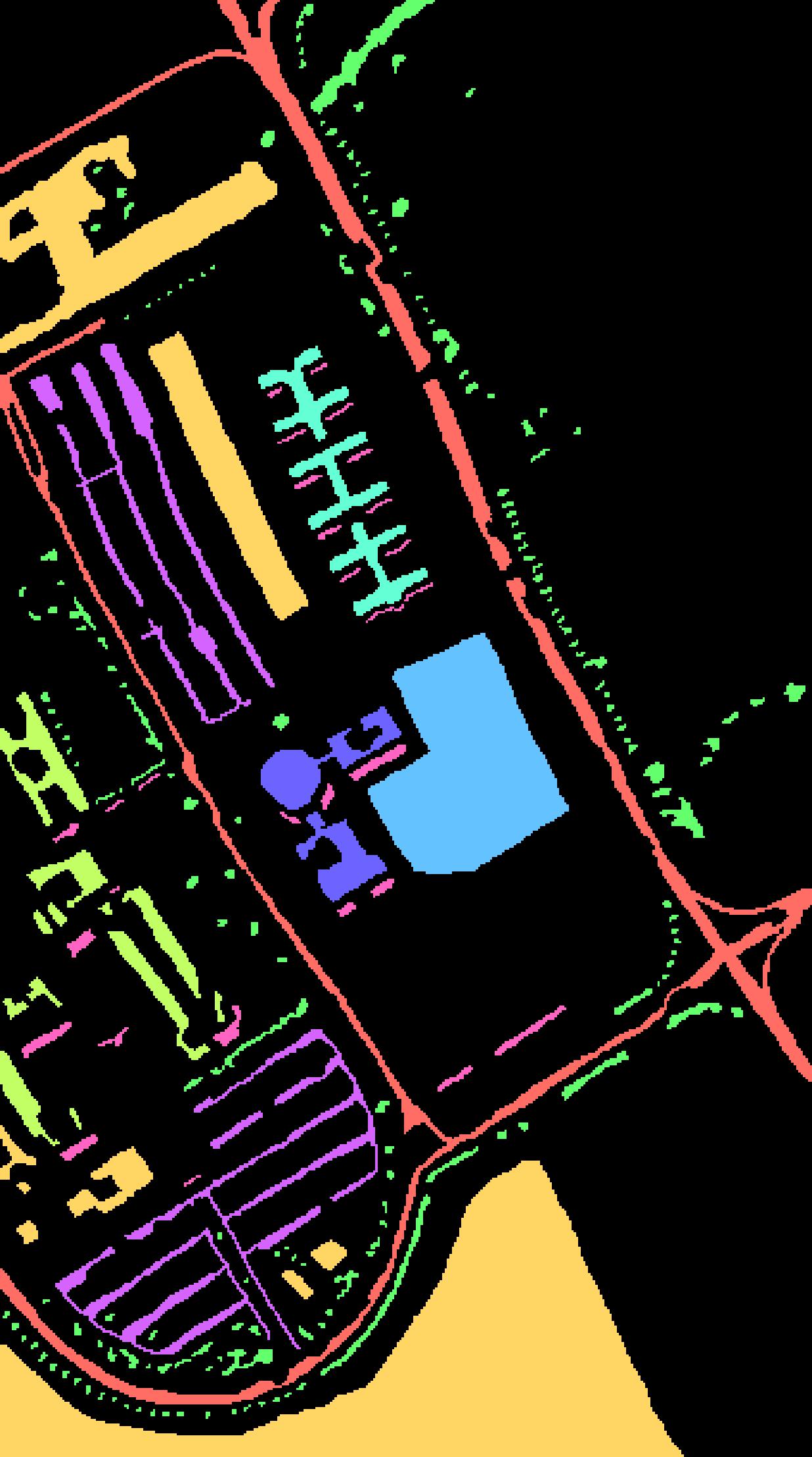}}
   \subfigure[]{\label{fig:d1_c}\includegraphics[width=0.16\linewidth]{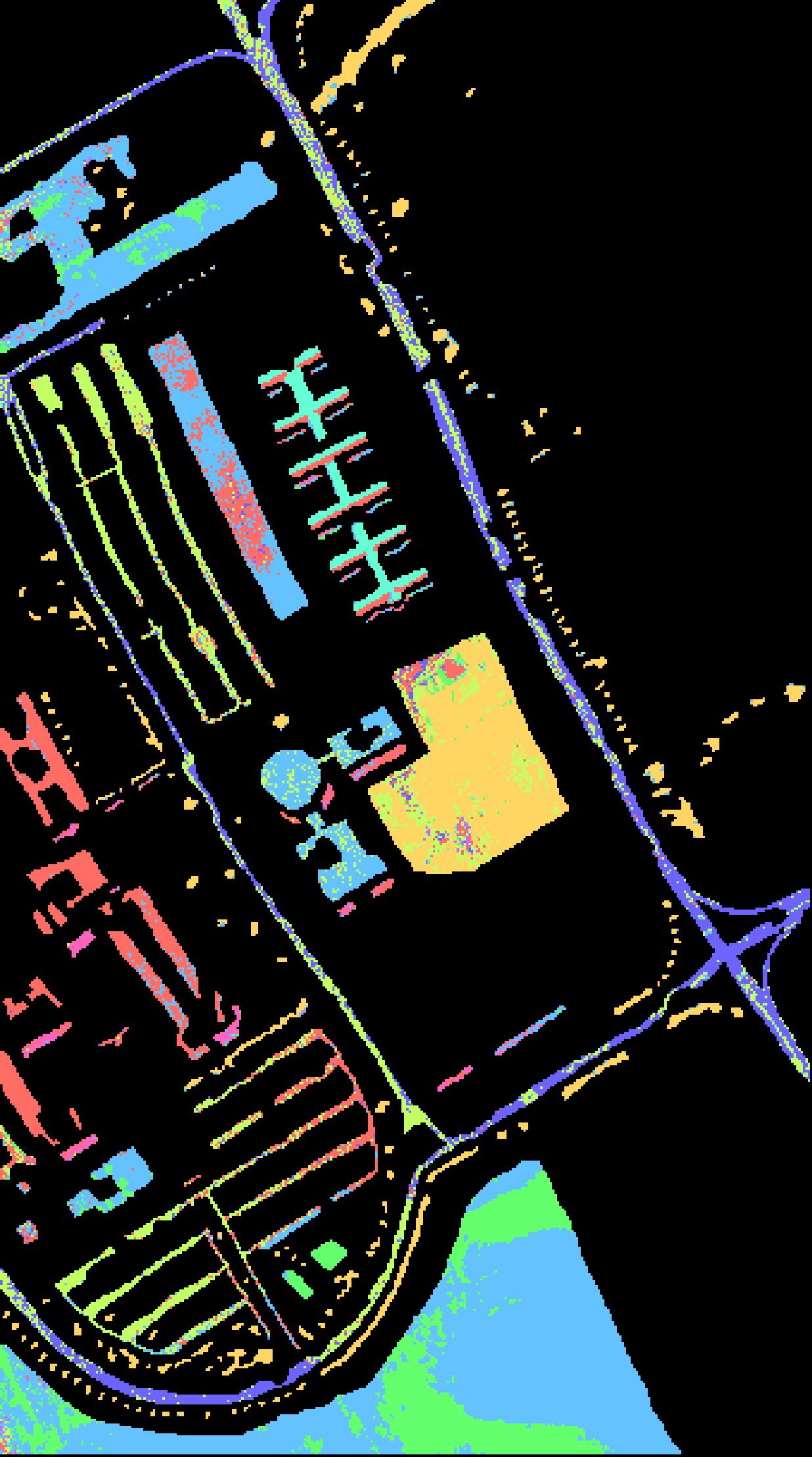}}
   \subfigure[]{\label{fig:d1_d}\includegraphics[width=0.16\linewidth]{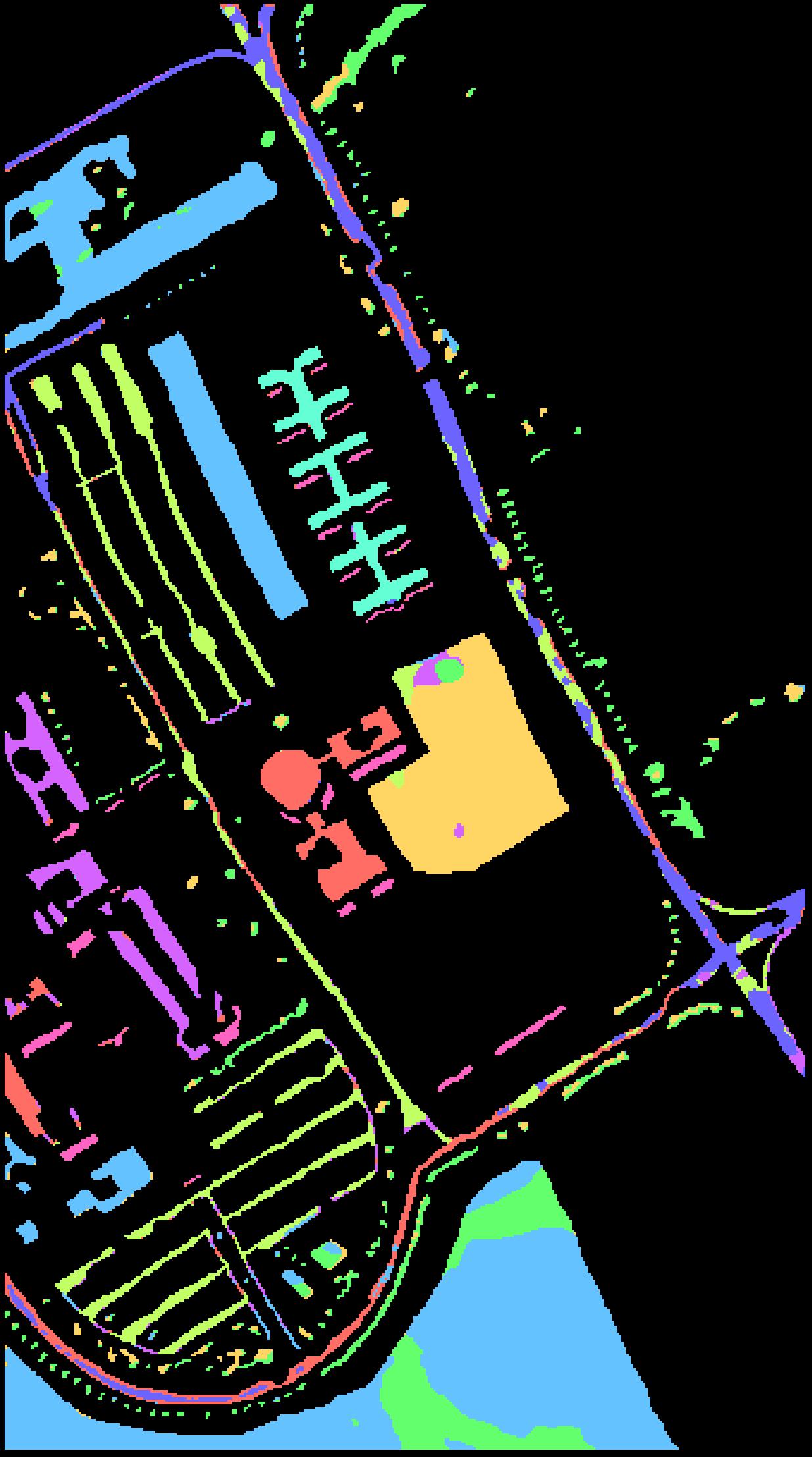}}
   \subfigure[]{\label{fig:d1_e}\includegraphics[width=0.16\linewidth]{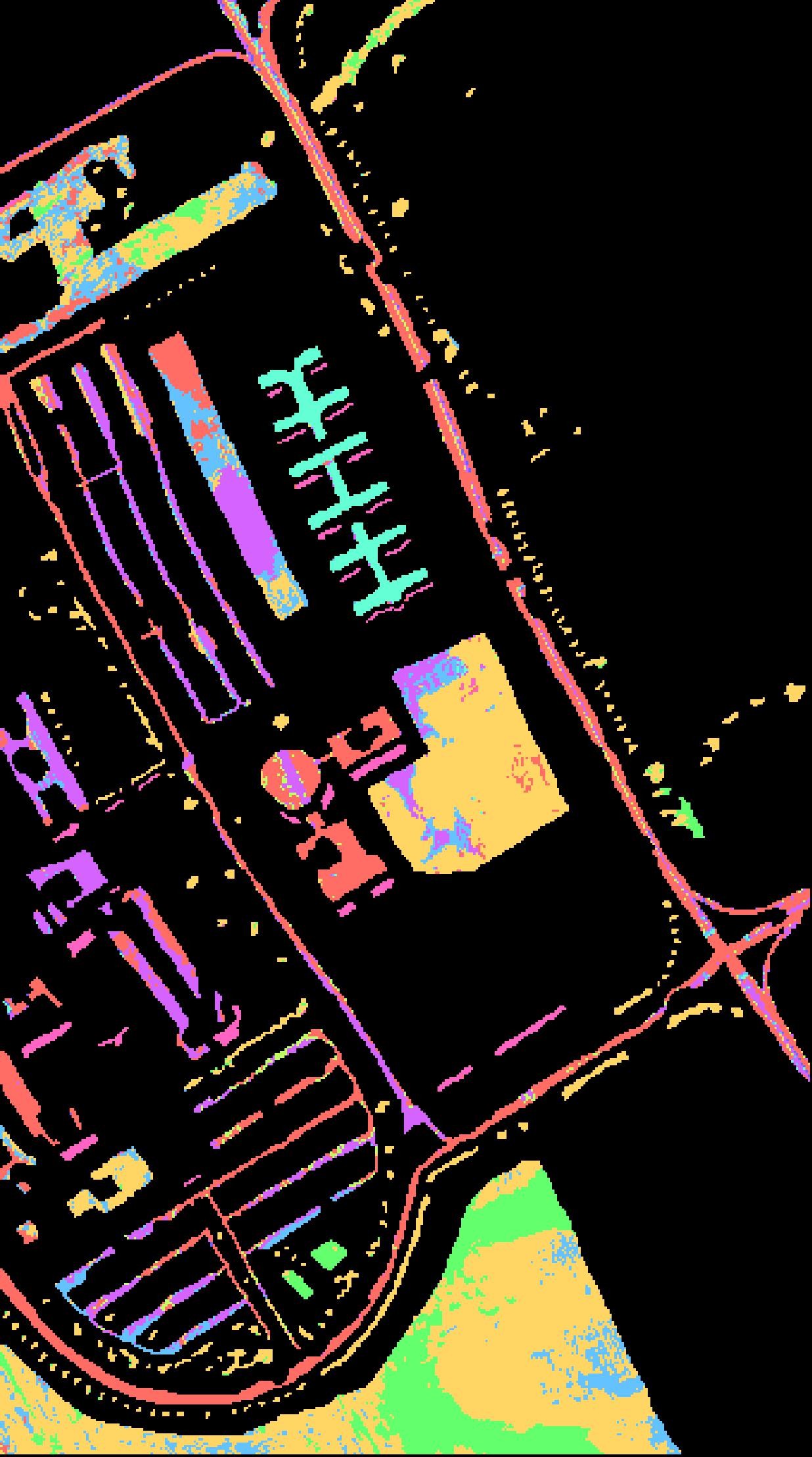}}
   \subfigure[]{\label{fig:d1_f}\includegraphics[width=0.16\linewidth]{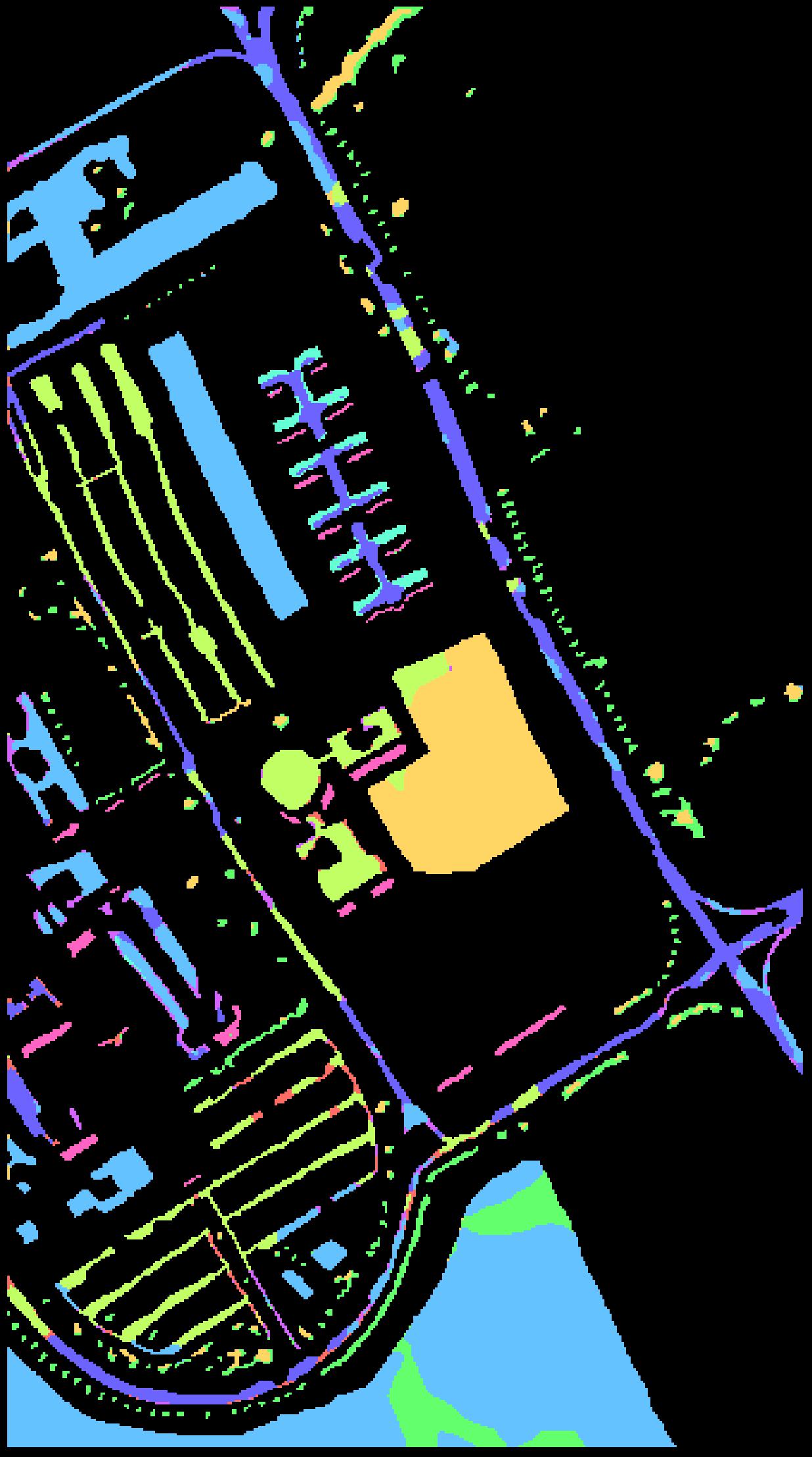}}
   \subfigure[]{\label{fig:d1_g}\includegraphics[width=0.16\linewidth]{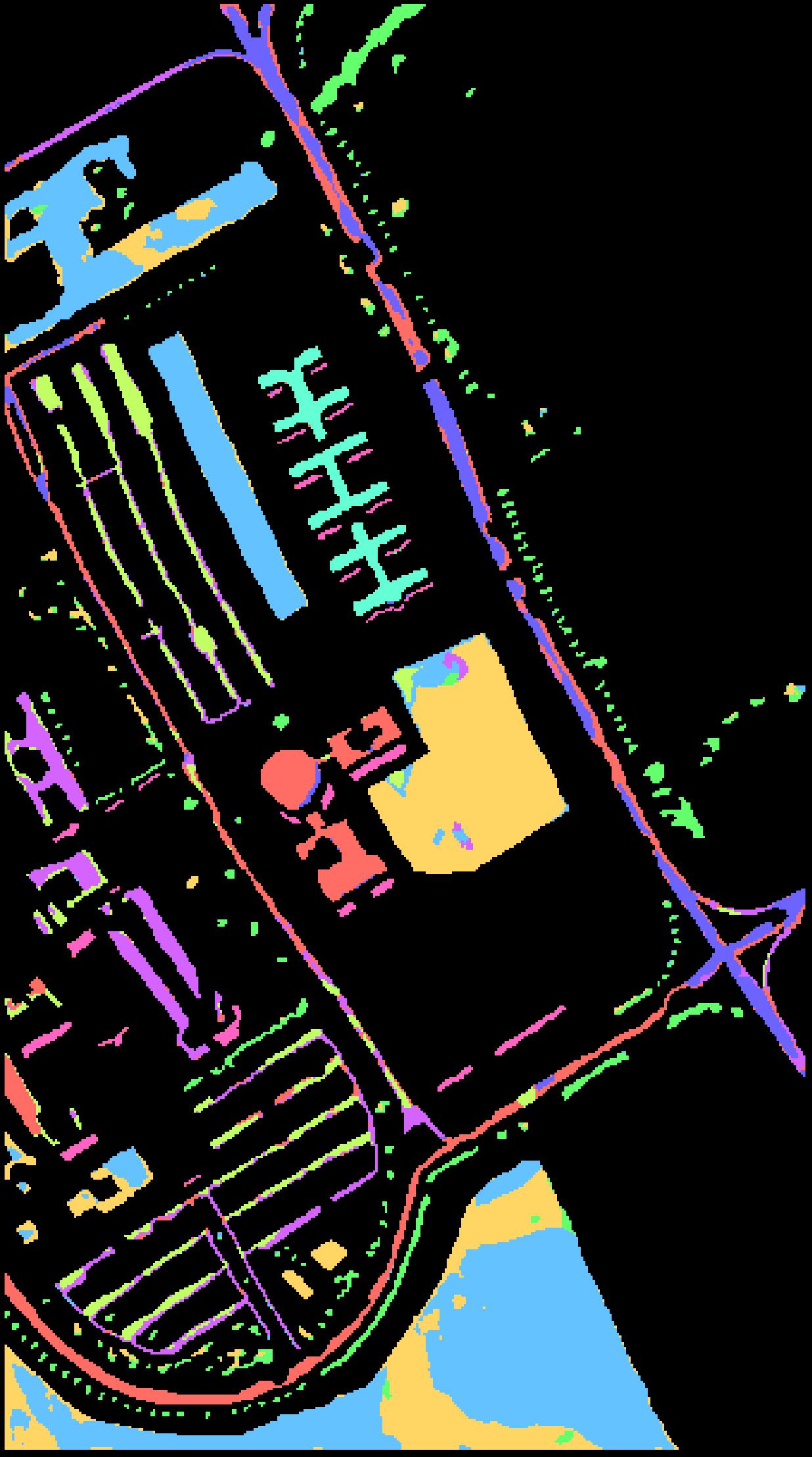}}
   \subfigure[]{\label{fig:d1_h}\includegraphics[width=0.16\linewidth]{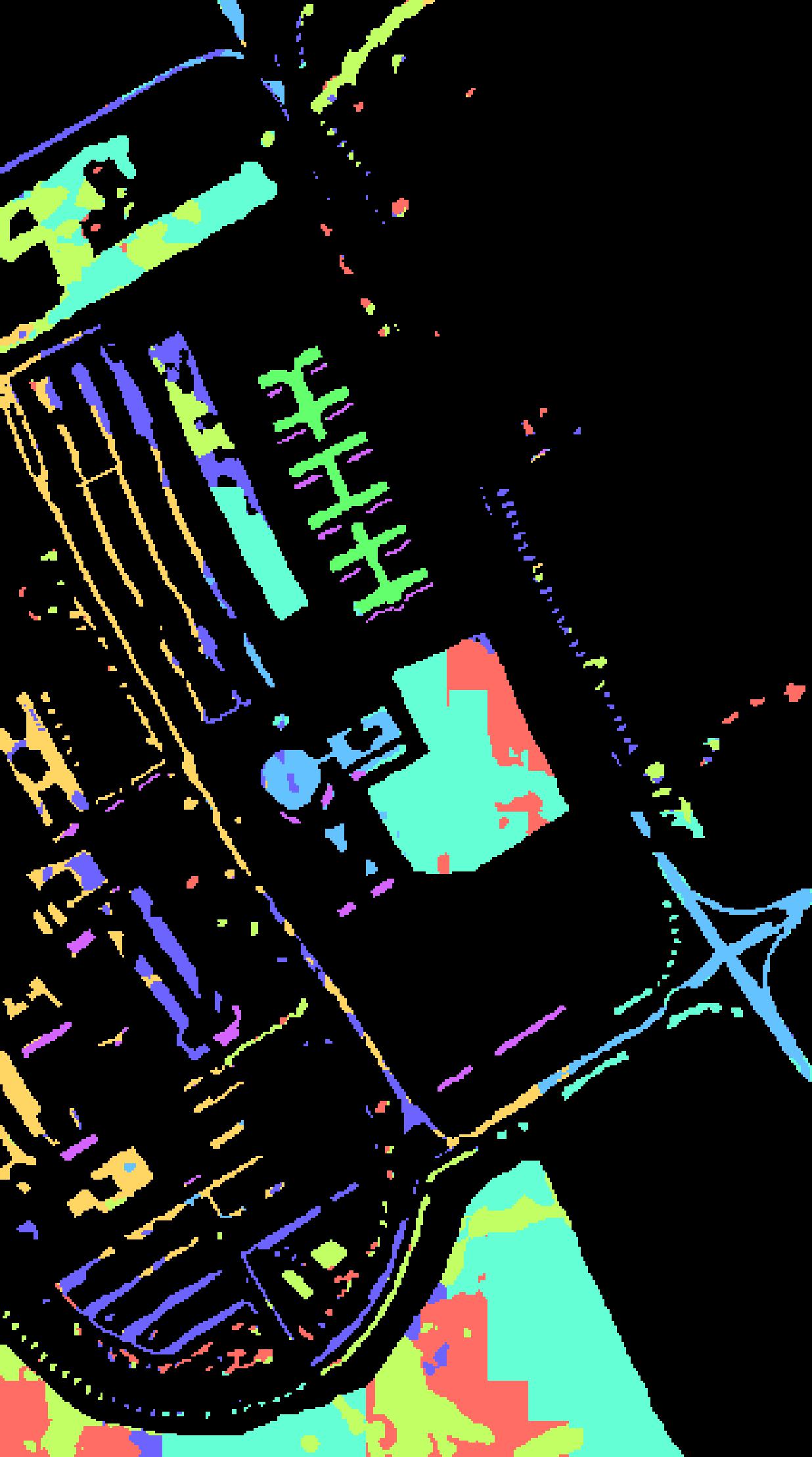}}
   \subfigure[]{\label{fig:d1_i}\includegraphics[width=0.16\linewidth]{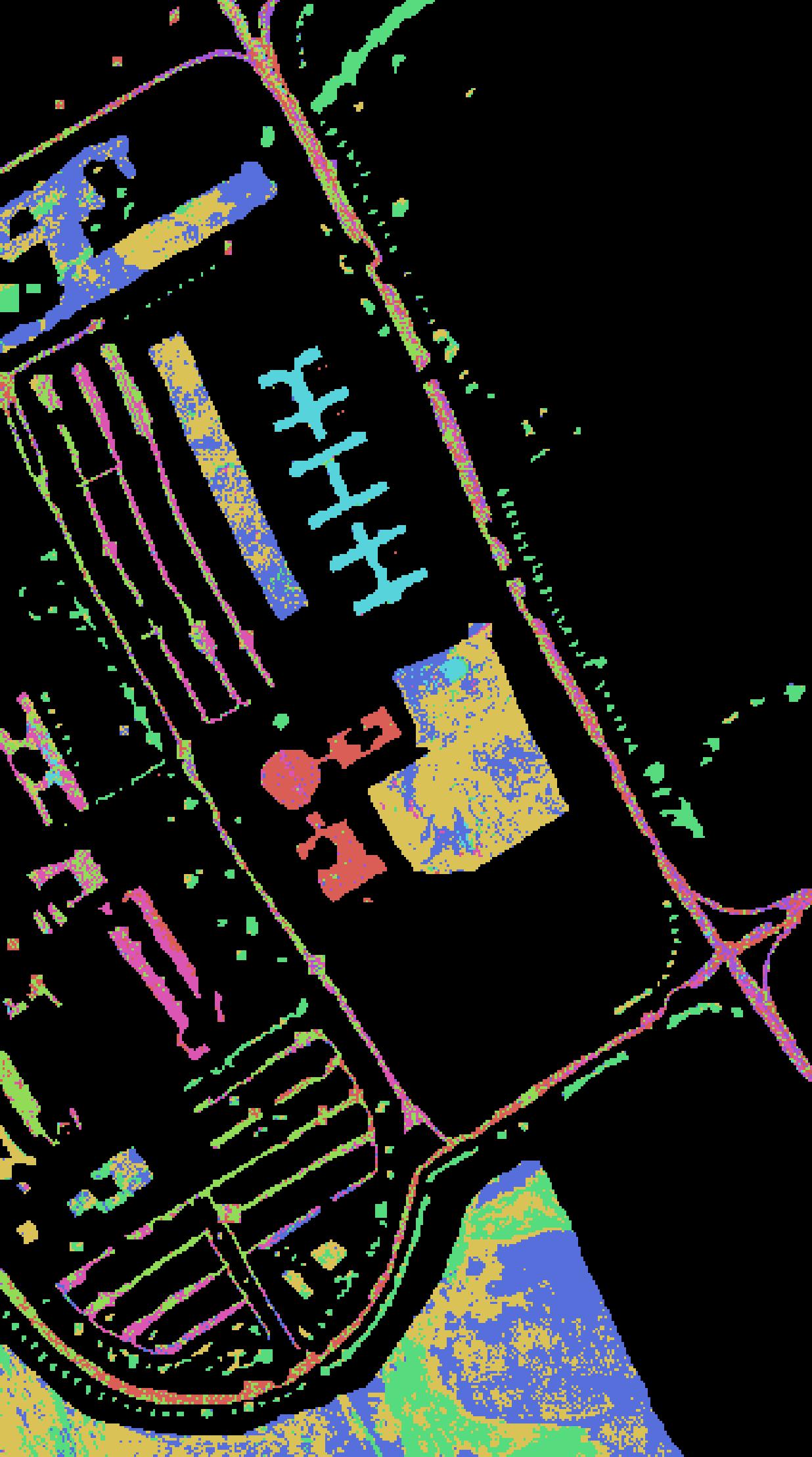}}
   \subfigure[]{\label{fig:d1_j}\includegraphics[width=0.16\linewidth]{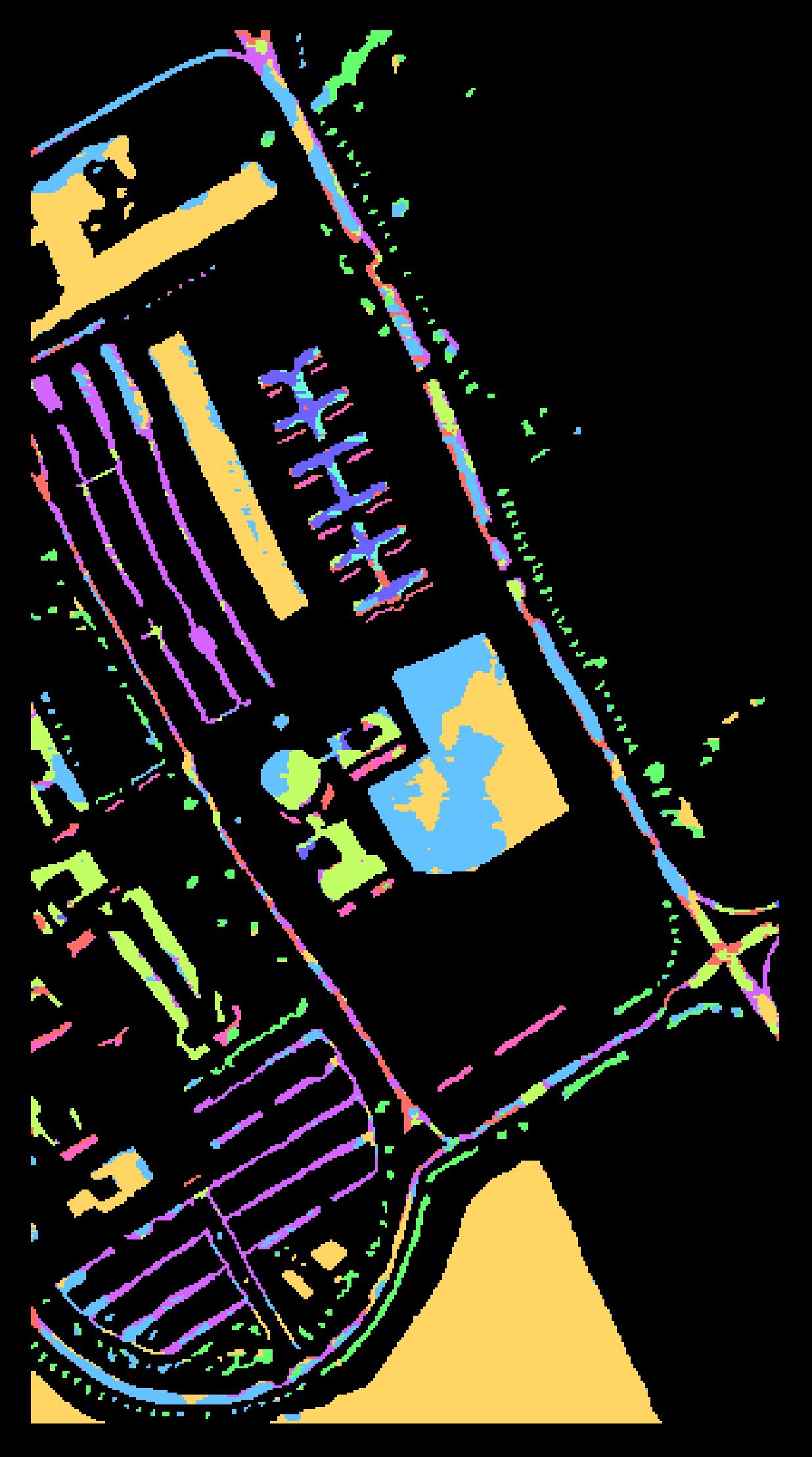}}
   \subfigure[]{\label{fig:d1_k}\includegraphics[width=0.16\linewidth]{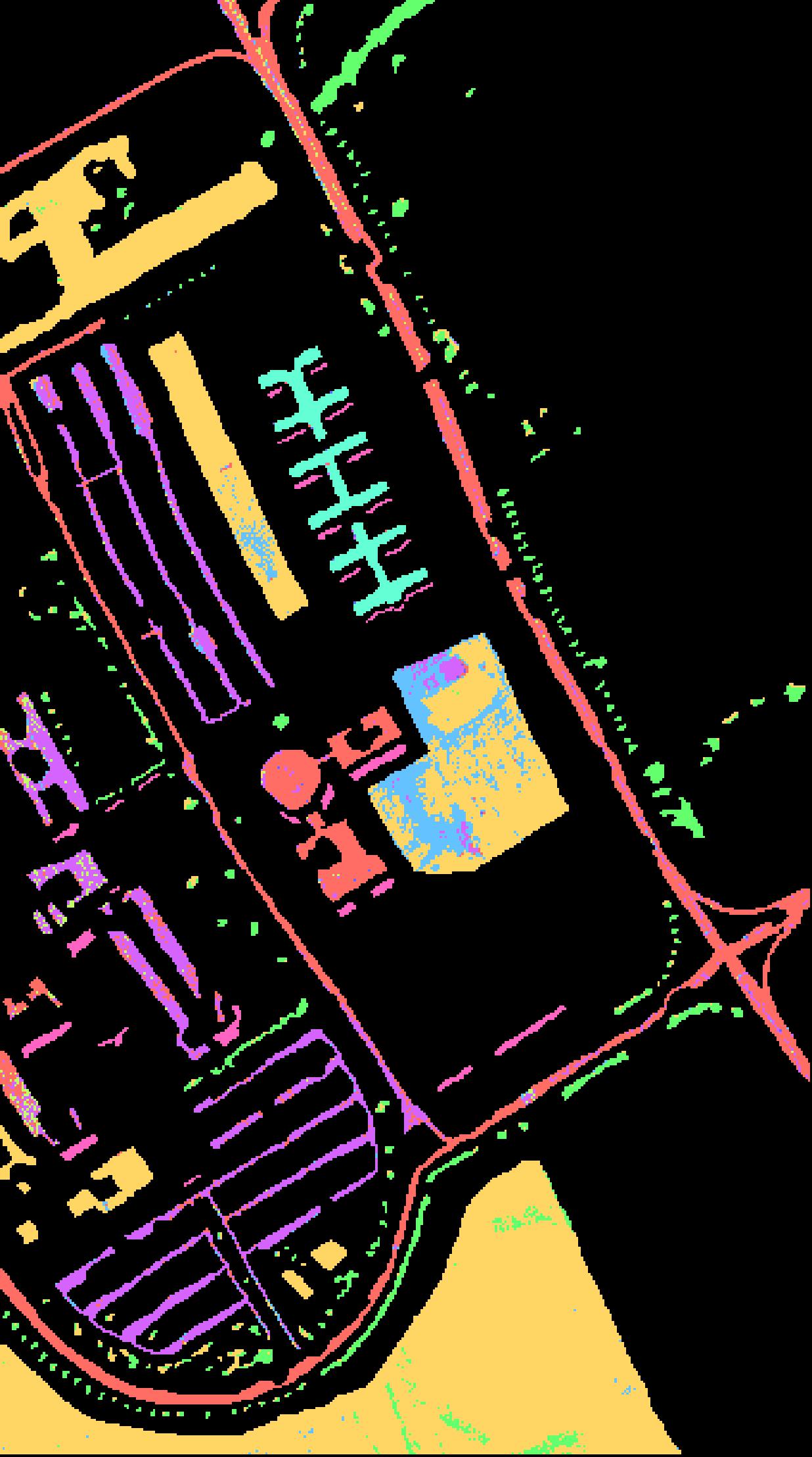}}
   \subfigure[]{\label{fig:d1_l}\includegraphics[width=0.16\linewidth]{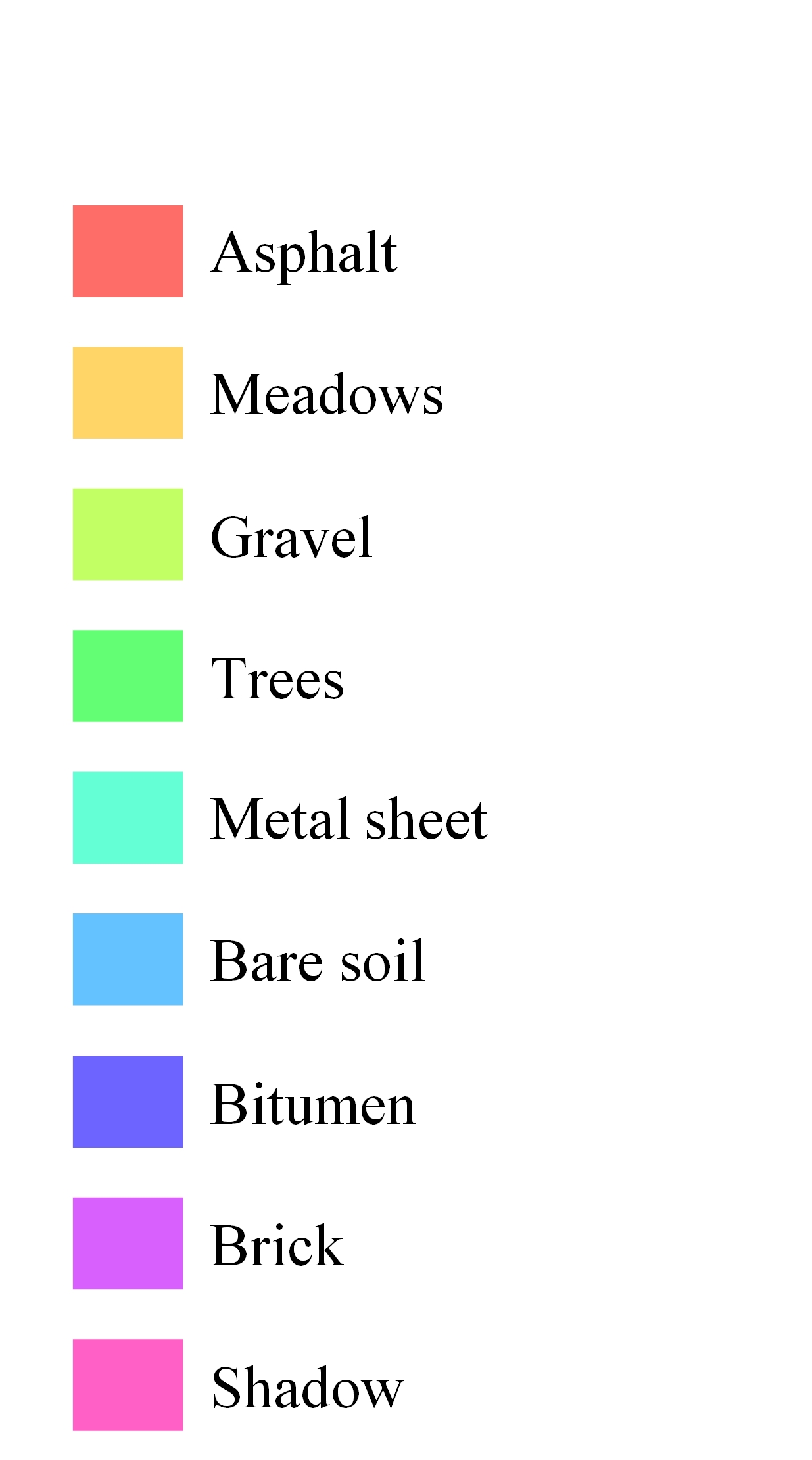}}
     \caption{Classification results on PaviaU dataset under FGSM adversarial attack. (a) The false color image; (b) Ground-truth; (c) Baseline; (d) 1-D CNN; (e) 3-D CNN; (f) HybridSN; (g) DRNN; (h) SpectralFormer; (i) WFCG; (j) SACNet; (k) MSSA; (l) Map color.}
  \label{fig:D1}
\end{figure*}
As for MSSA, it achieves OAs greater than 70\% for all the datasets under different attack methods, which significantly exceeds the comparing DL hyperspectral classifiers. 
This convincingly validates that, MSSA succeed in promoting the intrinsic resistibility of DL framework. 
The reason is that, MSSA has filtrated most of adversarial perturbations with masked sequence attention learning module during the data input stage. 
Meanwhile, the global associative representation module helps to generate more robust classification results by building global pixel-wise combinations. 
Unfortunately, since only a naive fully connected network (baseline) is utilized as the classifier, MSSA fails to achieve the best performance on clean datasets.
But MSSA dramatically improves the classification performances of baseline, which proves that it is also a powerful feature extractor.
\par 
\begin{table}[!t]
  \renewcommand{\arraystretch}{1.25}
  \scriptsize
  \begin{center}
  \caption{Quantitative performances of comparing adversarial defense strategies in PaviaU dataset.}
\label{table:6}
\begin{tabular}{| c | c | c | c | c | c | c |}
\hline
{Method}     & {Metrics} & {YOPO}  & {TE} & {ALS} & {\cellcolor{lightgray!50} MSSA}\\
\hline\hline
\multirow{3}{*}{FGSM}& {OA(\%)}   & 69.49±15.9 & 62.25±6.27 & 64.7±2.6 & \textbf{81.87±4.85} \\
& {AA(\%)}    & 65.08±6.13 & 41.04±3.98 & 59.37±15.8 & \textbf{79.04±5.91} \\
& {$\kappa$ (\%)}  & 60.45±3.14 & 49.03±8.08 & 55.08±10.1 & \textbf{75.13±1.93}  \\
 \hline\hline
\multirow{3}{*}{BIM}& {OA(\%)}   & 37.54±9.47 & 60.27±2.25 & 54.36±0.88 & \textbf{76.93±1.13} \\ 
& {AA(\%)}   & 40.1±16.49 & 39.34±7.46 & 53.76±18.1 & \textbf{69.16±4.62} \\
& {$\kappa$ (\%)}   & 23.8±15.26 & 46.47±9.05 & 43.1±5.05 & \textbf{68.53±1.81} \\
 \hline\hline
 \multirow{3}{*}{DeepFool}& {OA(\%)}  & 13.74±13.3 & 57.57±11.9 & 47.98±0.87 & \textbf{73.11±3.04} \\
 & {AA(\%)}   & 29.4±1.55 & 37.42±5.81 & 43.95±1.28 & \textbf{65.18±2.67} \\
 & {$\kappa$ (\%)}  & 3.7±16.16 & 43.06±9.01 & 31.7±16.14 & \textbf{63.2±1.06} \\
\hline
\end{tabular}    
\begin{tablenotes}
  \item The \textbf{bold entries} represent the best performance in each row.
\end{tablenotes}
\end{center}
\end{table}
To demonstrate the classification results from the visual aspect, Fig. \ref{fig:D1} illustrates the reference maps about aforementioned methods on PaviaU dataset with FGSM algorithm. 
It can be observed that all the comparing methods have been fooled by FGSM algorithm to the varying degrees.
The classification maps of baseline looks like a mess. 
In terms of other comparing methods, the 'Bare soil' category is always misclassified as 'Meadows' category. 
Meanwhile, they misclassify 'Bitumen' category into 'Asphalt' category as well. 
By contrast, MSSA seems to be more stable towards the adversarial attacks, because its classification map achieves the slightest difference with the ground-truth based on visual judgement. 
Similar phenomenona can also be found on other combinations of testing datasets and attack algorithms.
\subsection{Performances of  Defense Strategies}
In this section, to further validate the effectiveness of MSSA, three representative adversarial defense strategies would be employed for conducting comparative experiments. 
A brief description of each adversarial defense strategy is shown as follows: (1) You-only-Propagate-Once (YOPO) \cite{Zhanga2019}: an accelerated version of adversarial training with PGD algorithm under a $\ell_2$-norm constraint; (2) Thermometer encoding (TE) \cite{Buckman2018}: a defense strategy confronts adversarial attack by masking the gradient information of the DL models; (3) Adversarial label smoothing (ALS) \cite{Fu2020}: a defense strategy improves the robustness by using label smoothing skills.
\par
The quantitative comparison results are reported in Table \ref{table:6} to Table \ref{table:8}. 
According to these tables we can find that, YOPO achieves impressive defense performances to confront FGSM method but has limited impact under the attacks of BIM and DeepFool. 
This phenomenon indicates that adversarial training-based defense methods are of poor transfer capacities, since they can only take effect when the adversarial perturbations are generated by the same norm constraints. 
In terms of TE, it achieves almost identical and competitive performances with different attack methods over three datasets. 
However, the performances of TE on Honston2013 dataset are worse than other two datasets, where the best OA value is merely 40.3\%. 
Class-imbalance might be the reason for this abnormal situation, because most of pixels in Honston2013 datasets are background pixels. 
As for ALS, it works well under FGSM and BIM methods, which only employs gradient to conduct attacks.
Nevertheless, a least 7\% performance decrease takes place if ALS is employed to counter DeepFool method. 
All of above results confirm that, the common adversarial defense strategies have limited performances in real-world applications. 
\par 
By contrast, MSSA can be well-behaved regardless of the forms of attack methods the distributions of training datasets. 
Take the results of MSSA in Houston2013 dataset for example, MSSA gets an OA of 79.51\%, while the suboptimum performance is only 43.23\%.
What's more, MSSA can achieve an OA of more than 70\% over all the datasets and attack methods, which significantly outperforms the comparing adversarial defense methods. 
Finally, we also exploit the execution time as another quantitative criterion and the corresponding results are listed in Table \ref{table:9}.
It is obvious that ALS is the most efficient method for all the datasets owing to its simple defense process. 
MSSA could not perform well in efficiency aspect, since exploiting more spectral and spatial contextual information is pretty time-consuming. 
However, the time consumption of MSSA seems to be tolerable while MSSA has achieved the best defense performances. 
\subsection{Ablation Study}
In this section, we will demonstrate the validation of each module in the proposed method and investigate how the hyper-parameters affect the defense performances. 
\par
\begin{figure*}[!t]
  \centering
   \subfigure[]{\label{fig:d3_a}\includegraphics[width=0.24\linewidth]{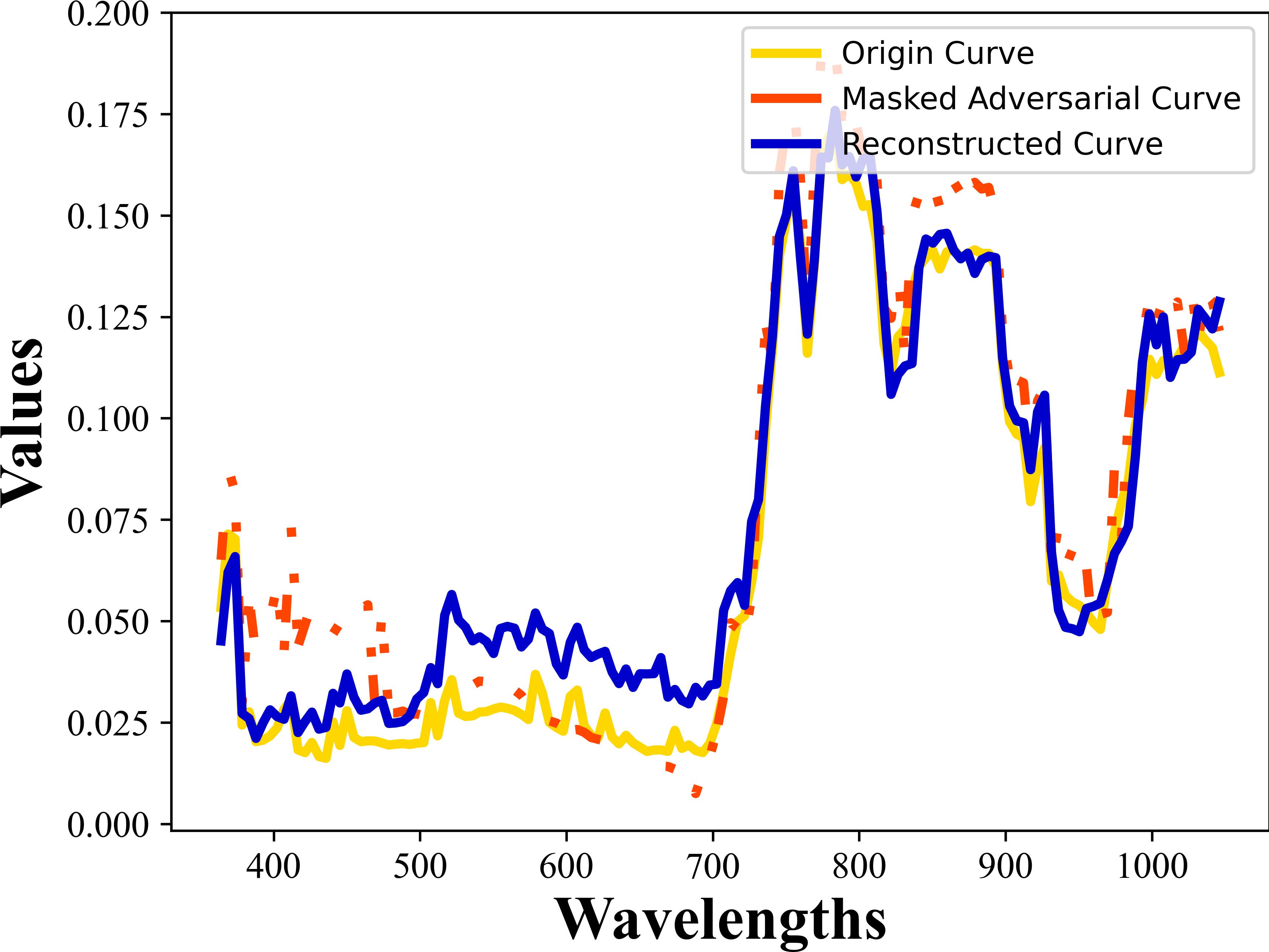}}
   \subfigure[]{\label{fig:d3_b}\includegraphics[width=0.24\linewidth]{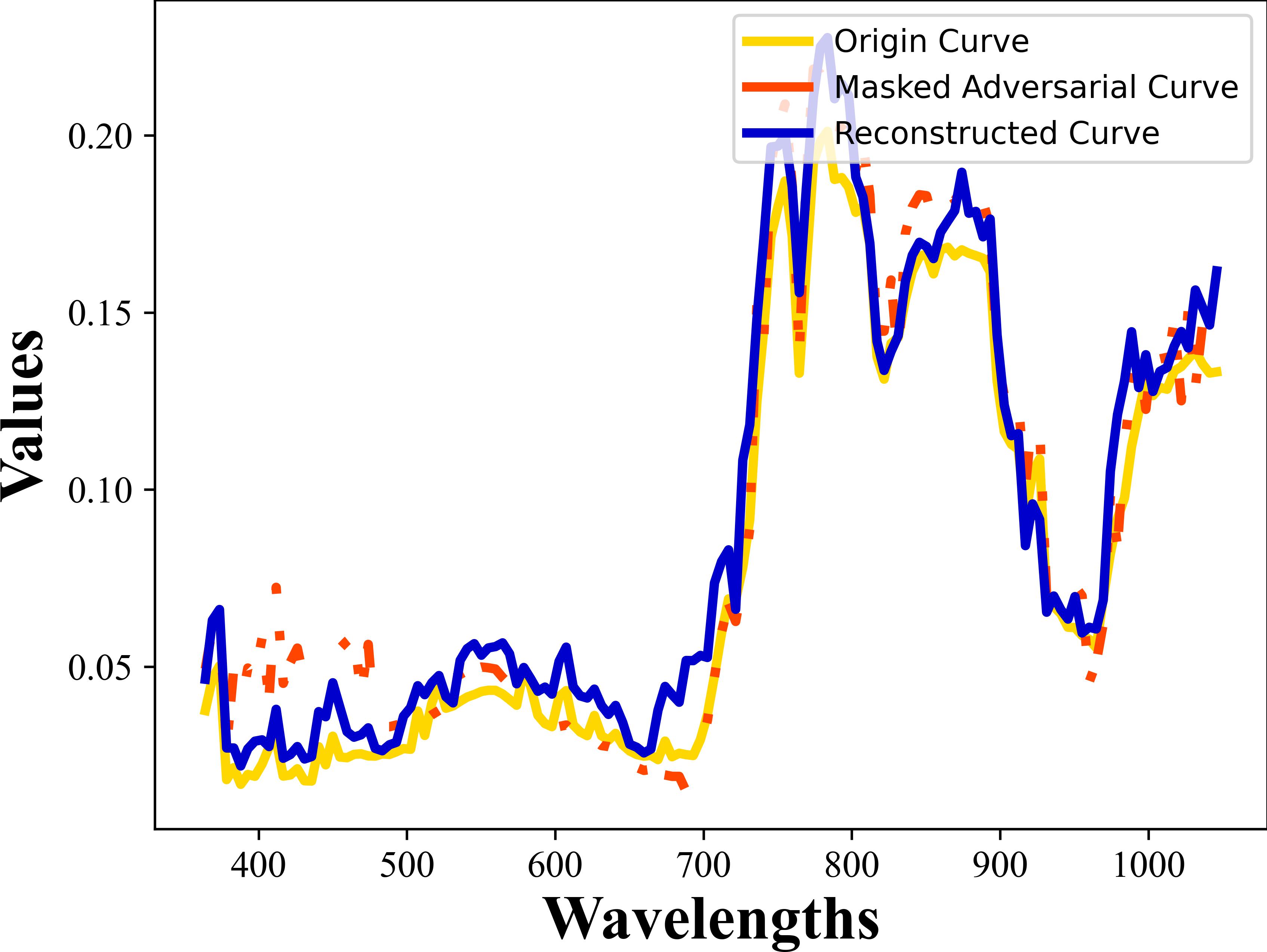}}
   \subfigure[]{\label{fig:d3_c}\includegraphics[width=0.24\linewidth]{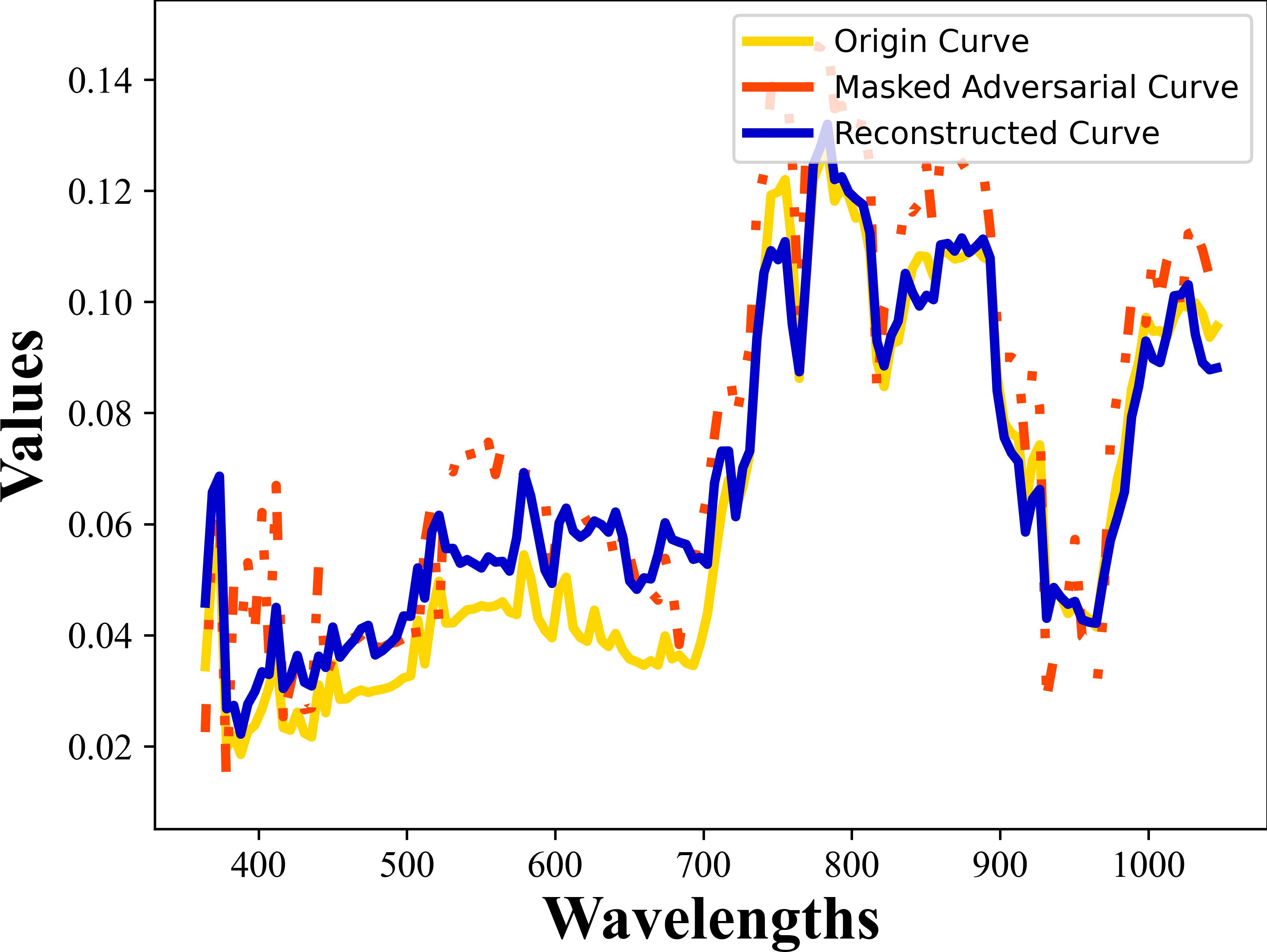}}
   \subfigure[]{\label{fig:d3_d}\includegraphics[width=0.24\linewidth]{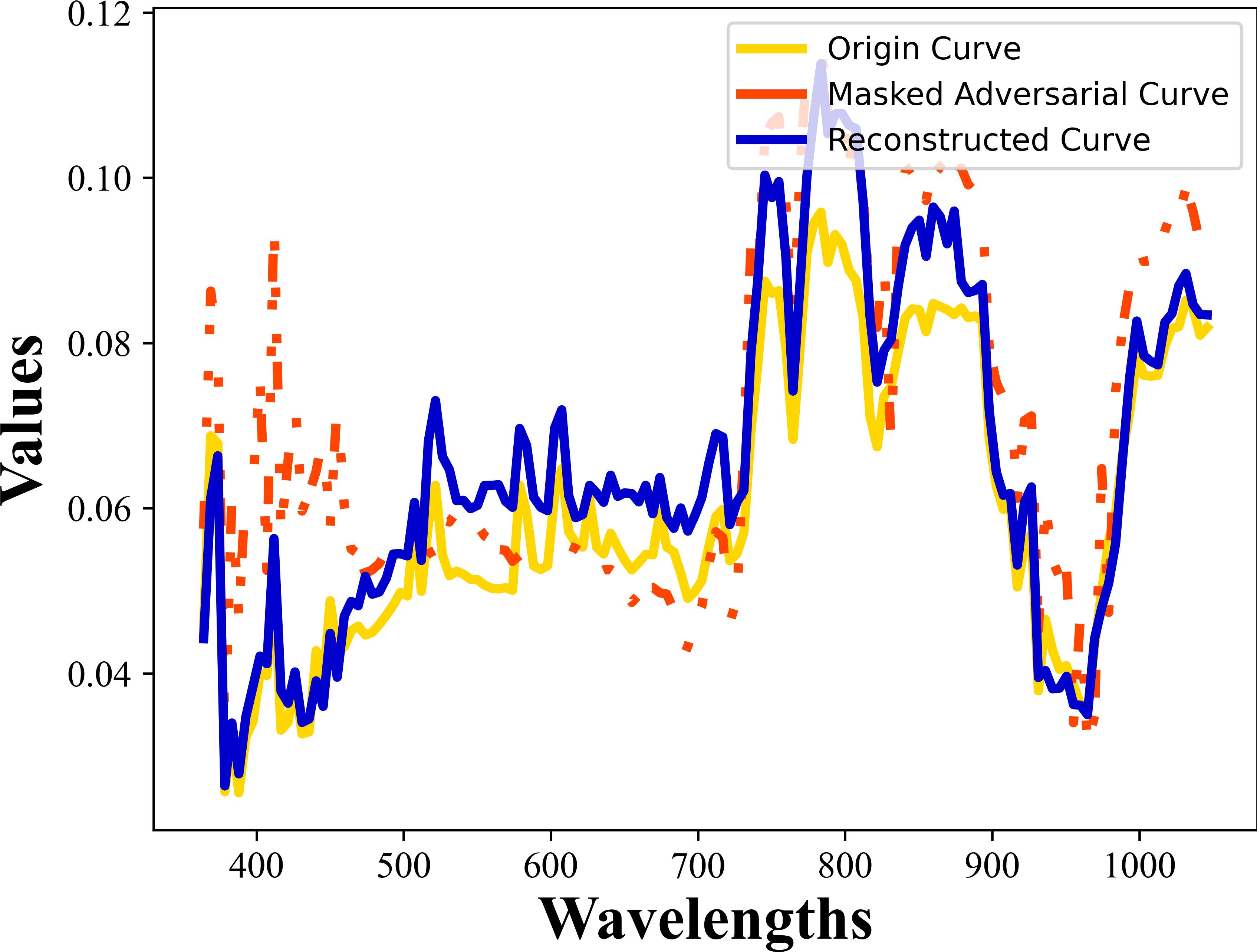}}
   \subfigure[]{\label{fig:d3_e}\includegraphics[width=0.24\linewidth]{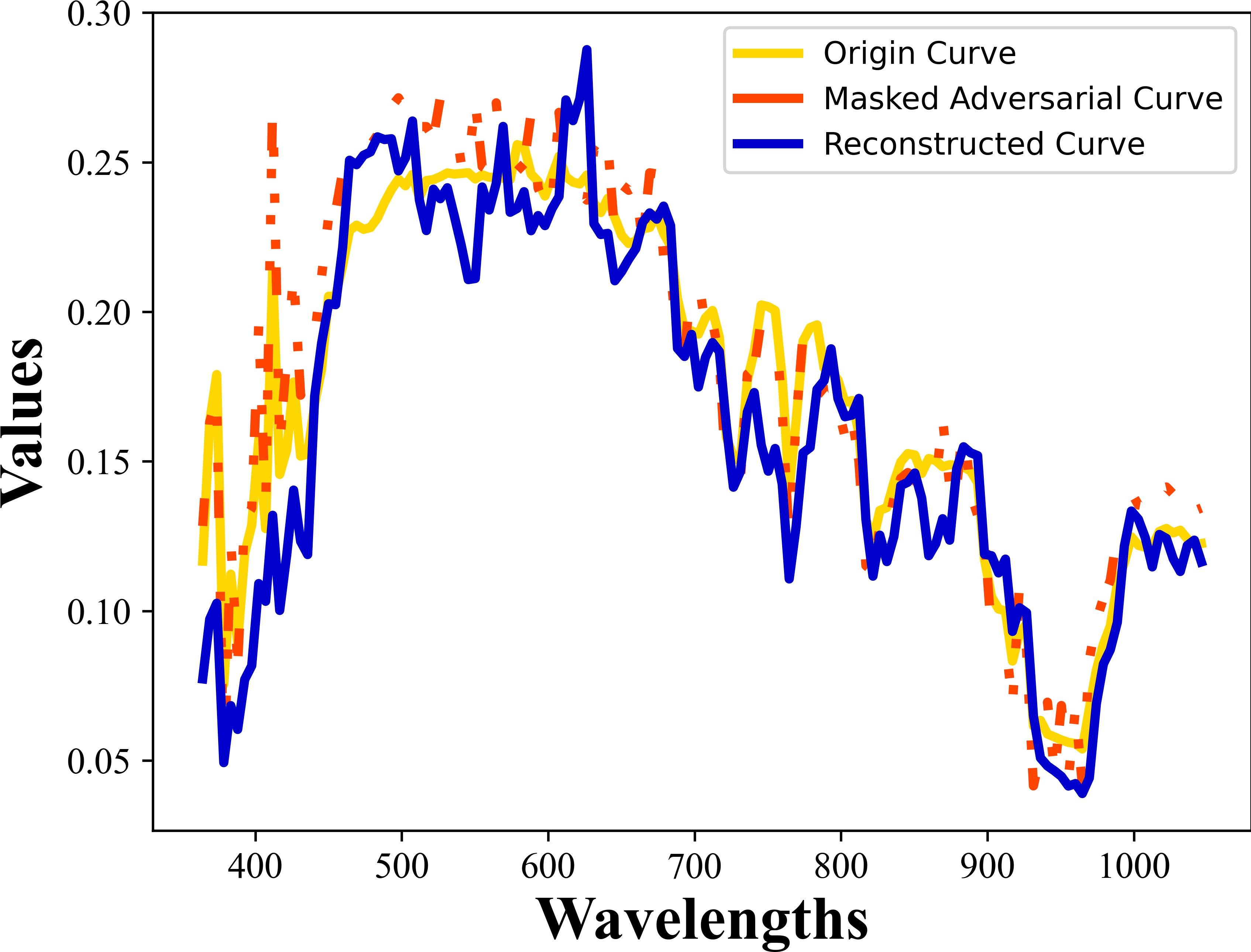}}
   \subfigure[]{\label{fig:d3_f}\includegraphics[width=0.24\linewidth]{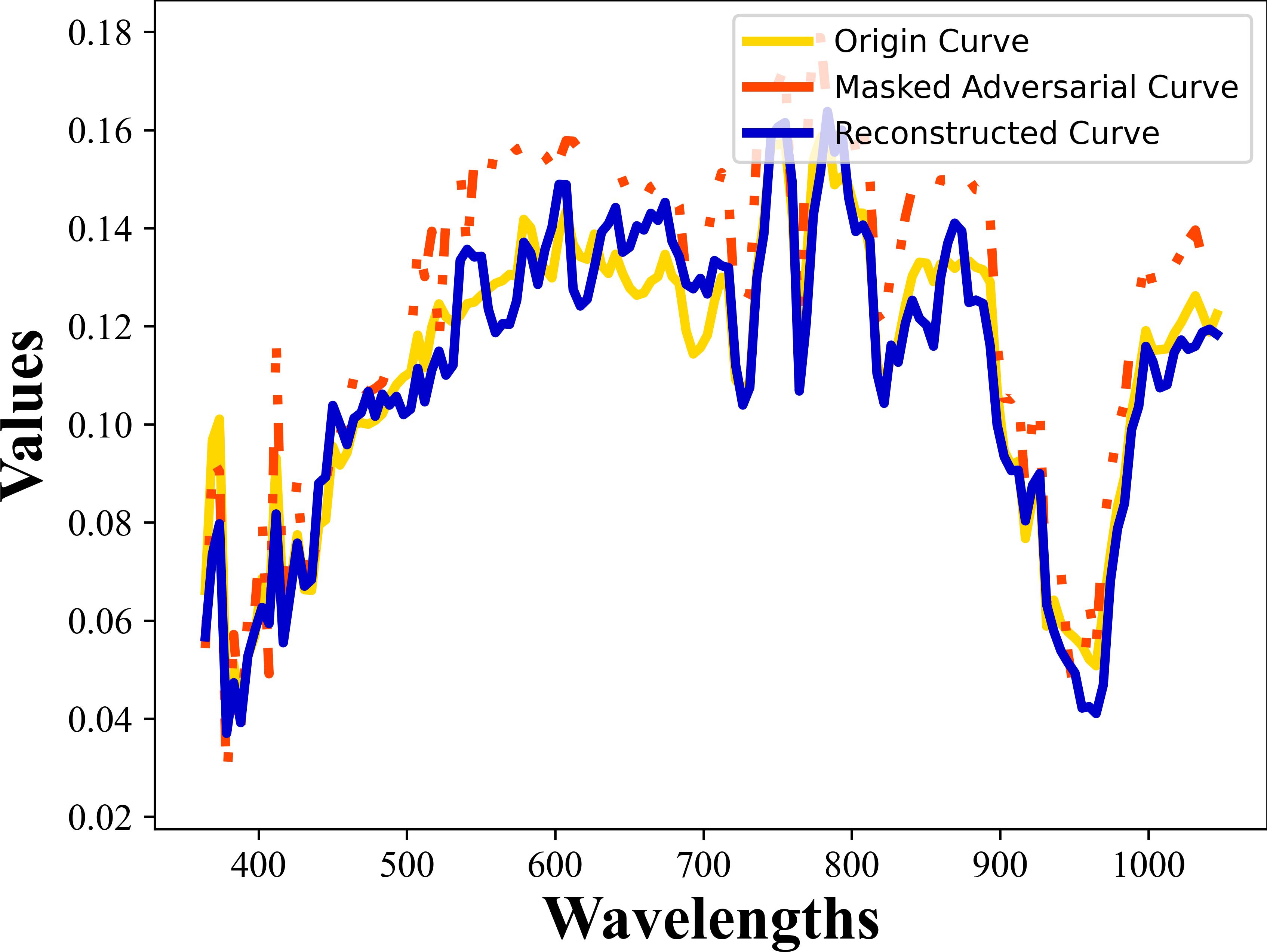}}
   \subfigure[]{\label{fig:d3_g}\includegraphics[width=0.24\linewidth]{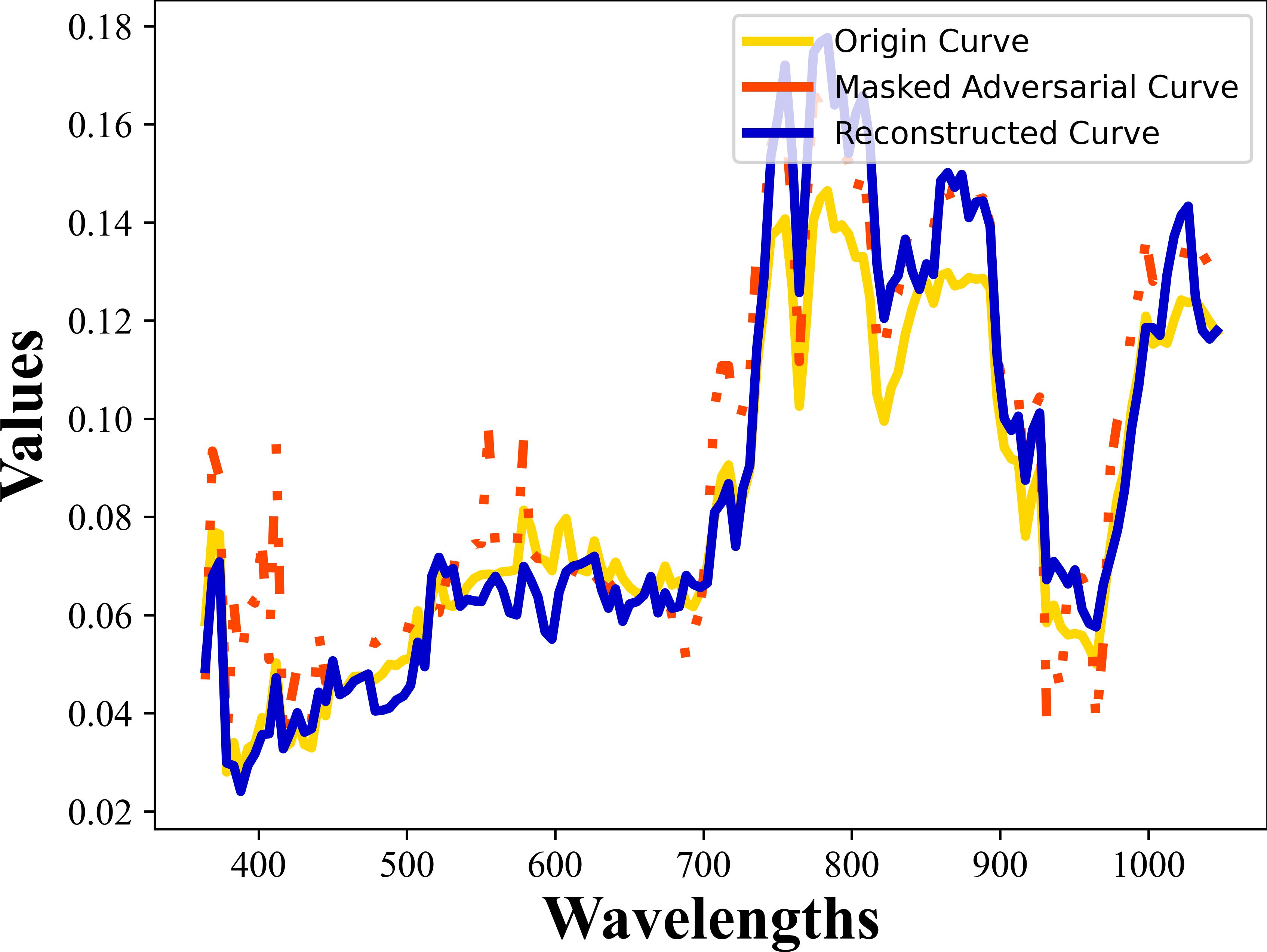}}
   \subfigure[]{\label{fig:d3_h}\includegraphics[width=0.24\linewidth]{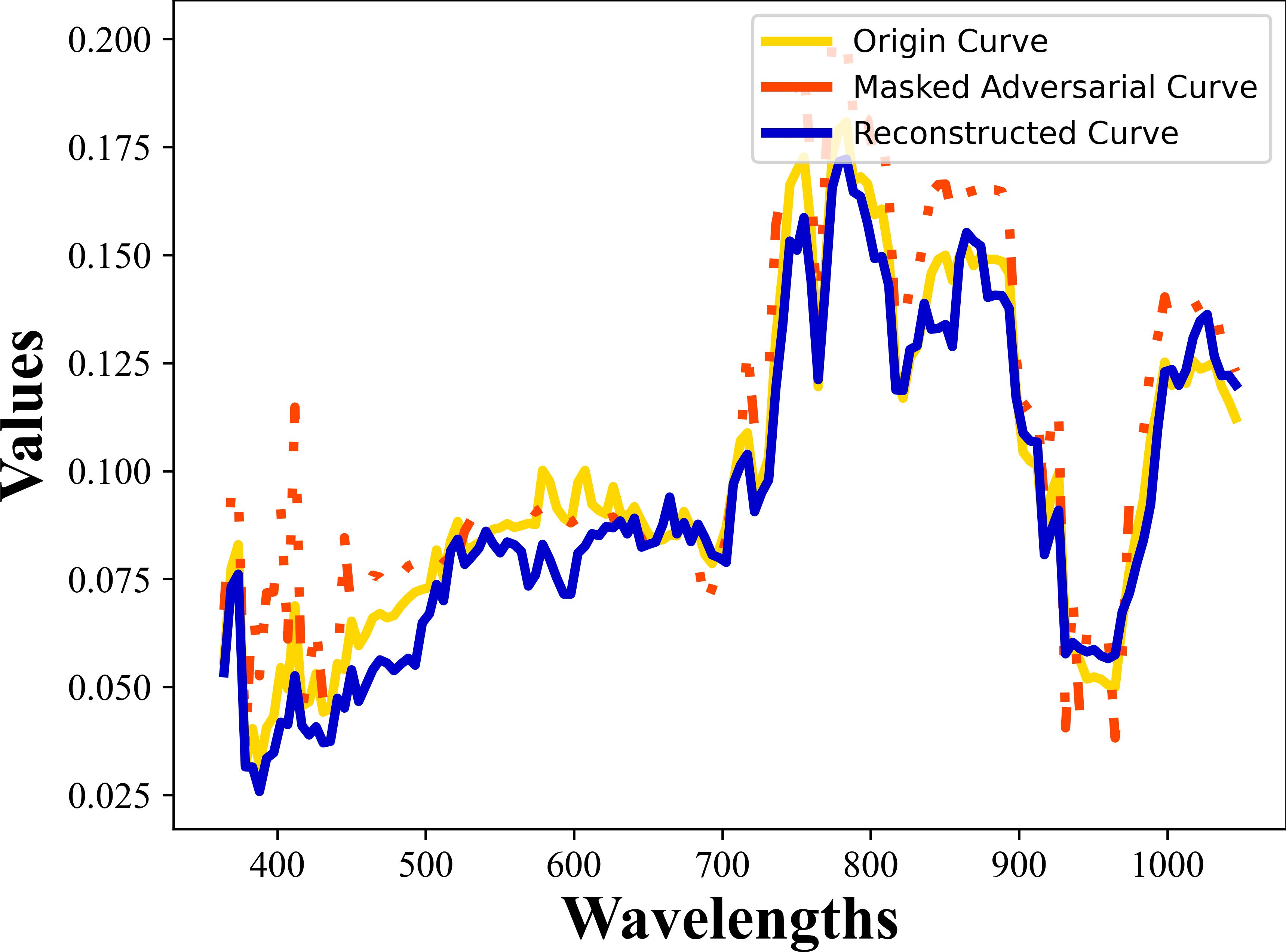}}
     \caption{The spectra reconstruction results underwater DeepFool attack with a masking ratio of 75\% in Houston2013 dataset. (a) Commercial; (b) Healthy grass; (c) Highway; (d) Parking Lot; (e) Stressed grass; (f) Road; (g) Running Track; (h) Water.}
  \label{fig:D3}
\end{figure*}

\begin{figure}[!t]
  \centering
   \subfigure[]{\label{fig:d4_a}\includegraphics[width=0.492\linewidth]{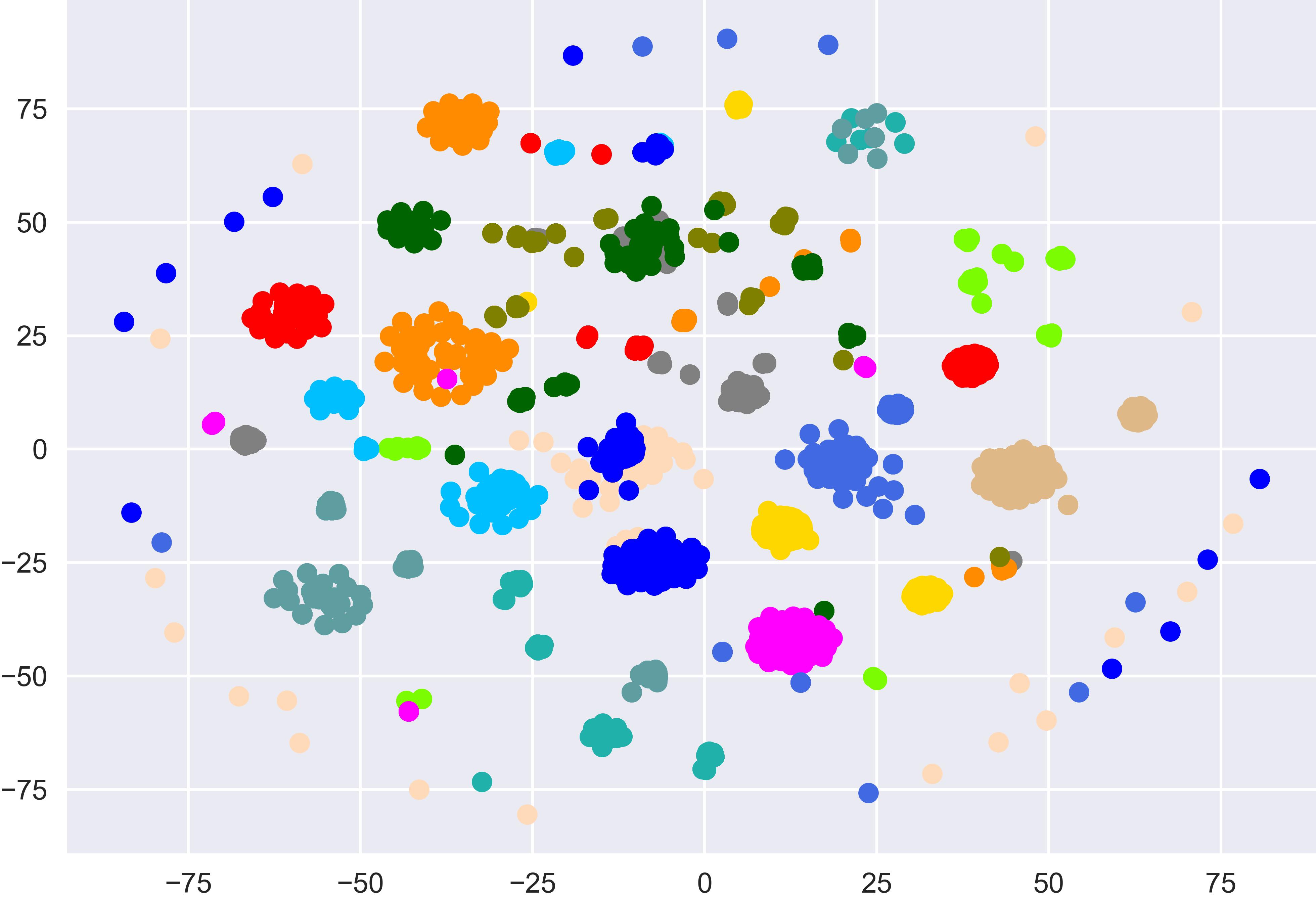}}
   \subfigure[]{\label{fig:d4_b}\includegraphics[width=0.492\linewidth]{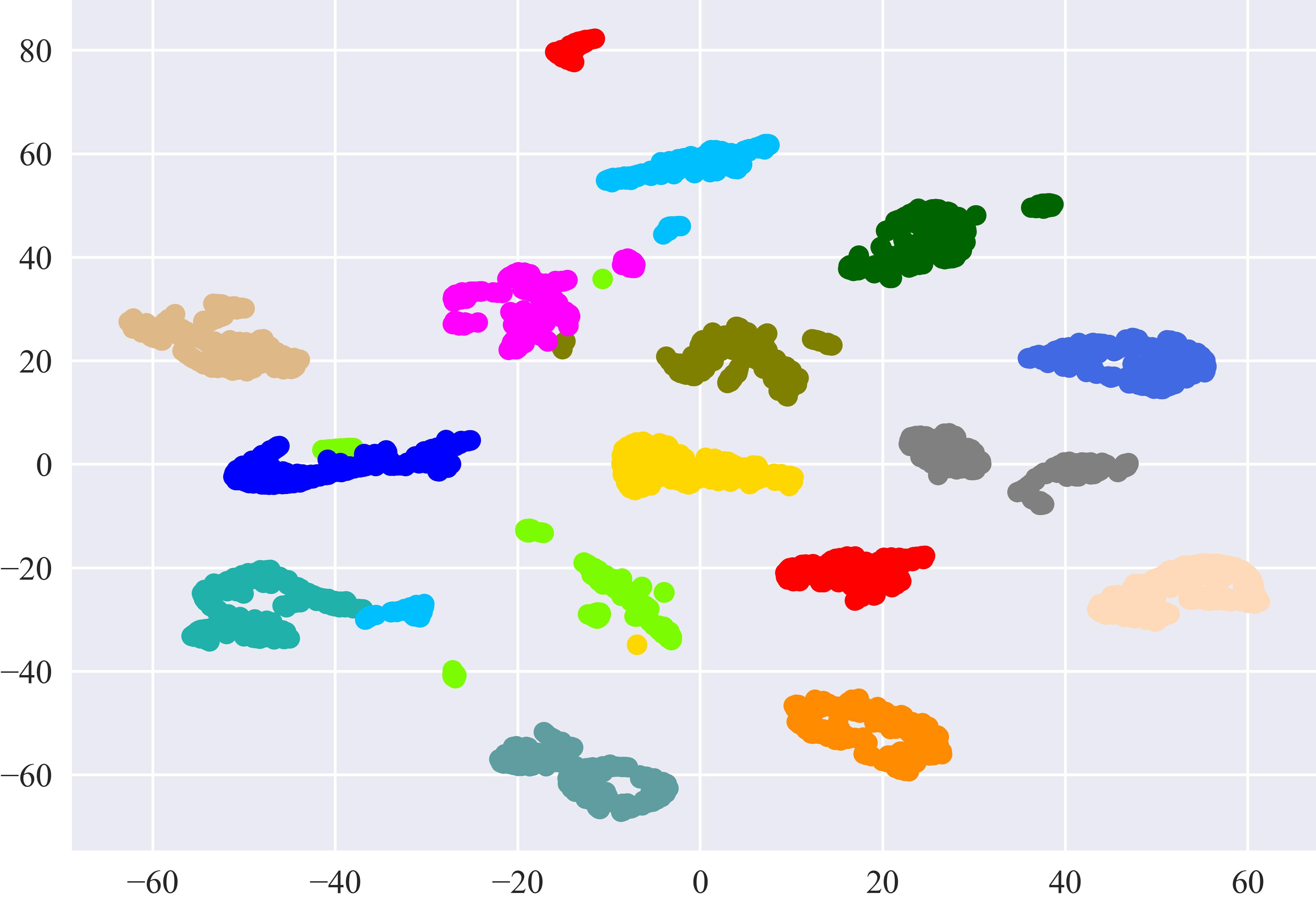}}
     \caption{ The correlation analysis based on T-SNE algorithm. (a) Without DGE module; (b) With DGE module.}
  \label{fig:D4}
\end{figure}

 
\textbf{1) Modules Validation Analysis}: The detailed defense performances with different modules are calculated and listed in Table \ref{table:10}. 
The average OA values of different datasets under all attack methods are employed as the evaluation metric to obtain an objective analysis result.
From this table we can infer that, the masked sequence attention learning (MSAL) module and DGE module dramatically improve the robustness of baseline and also contribute a lot to its classification performances. 
Furthermore, the MSAL module has a greater impact on the resistibility towards adversarial attacks while the DGE module is more conducive to the classification performances. 
Besides, Fig. \ref{fig:D3} and Fig. \ref{fig:D4} are exploited to visualize the effect of these two modules. 
In Fig. \ref{fig:D3}, the MSAL module succeeds in recovering the clean spectrums based on several masked adversarial spectrums under a masking ratio of 75\%. 
In Fig. \ref{fig:D4}, 200 pixels of each category in Houston2013 dataset are sampled to execute correlation analysis based on t-SNE algorithm \cite{Maaten2008}. 
The clustering results with DGE module are more compact and larger classification boundaries are achievable, which are beneficial to both the classification performances and model robustness. 
\par
\begin{figure}[!t]
  \centering
   \subfigure[]{\label{fig:d5_a}\includegraphics[width=0.49\linewidth]{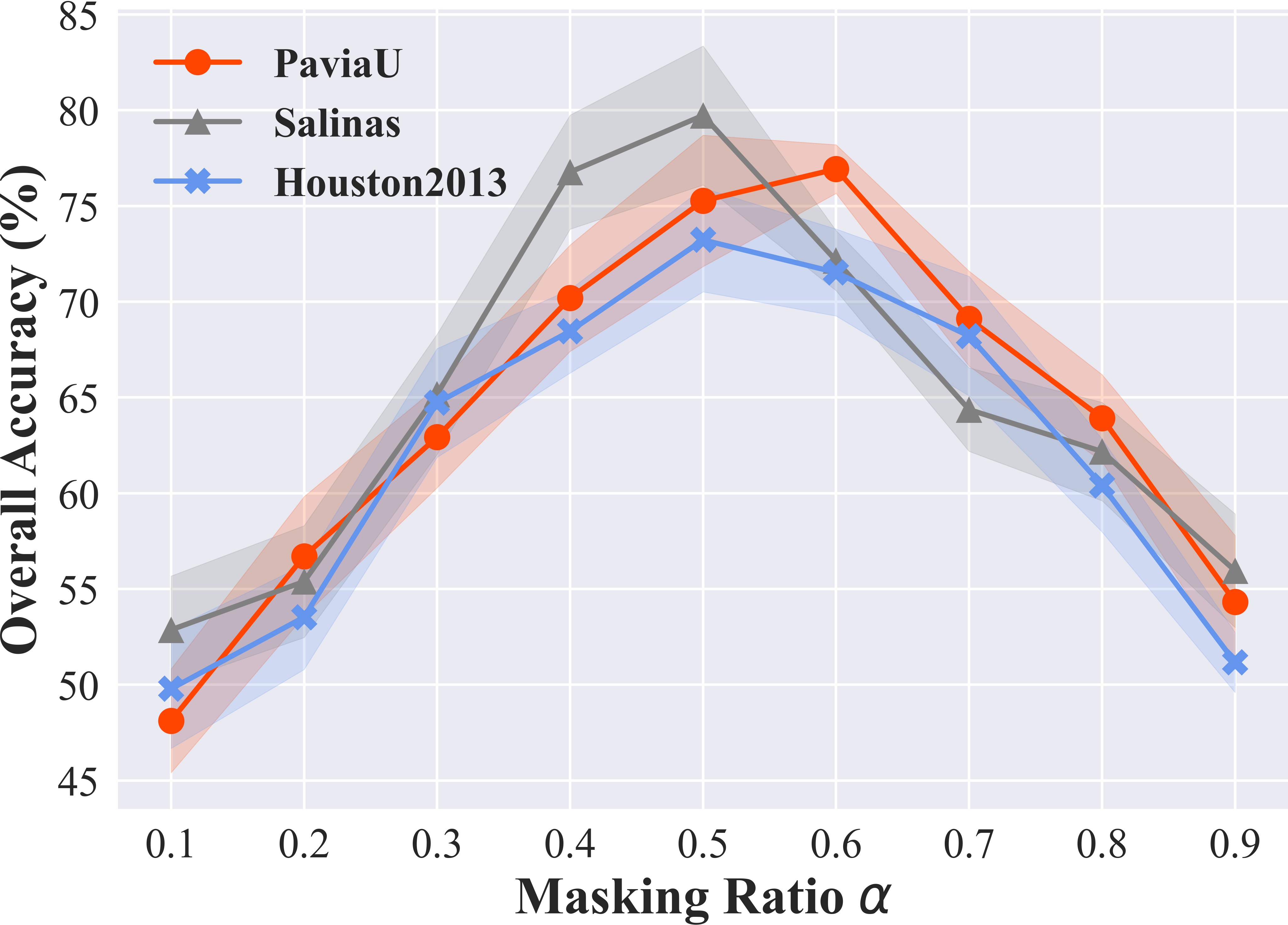}}
   \subfigure[]{\label{fig:d5_b}\includegraphics[width=0.49\linewidth]{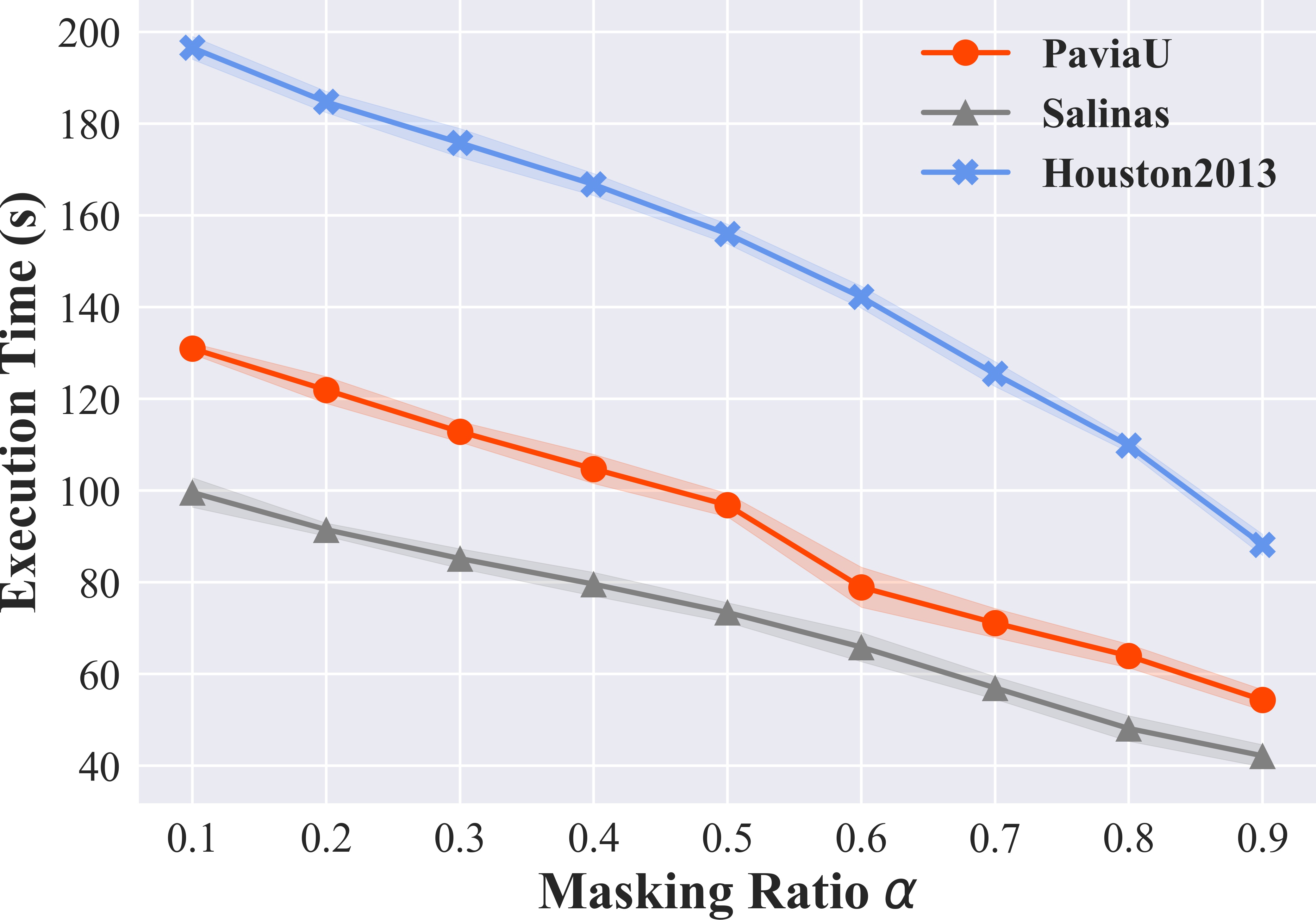}}
     \caption{The hyper-parameters sensitivity analysis results of the masking ratio $\alpha$. (a) Effectiveness; (b) Efficiency.}
  \label{fig:D5}
\end{figure}

\begin{figure}[!t]
  \centering
   \subfigure[]{\label{fig:d6_a}\includegraphics[width=0.49\linewidth]{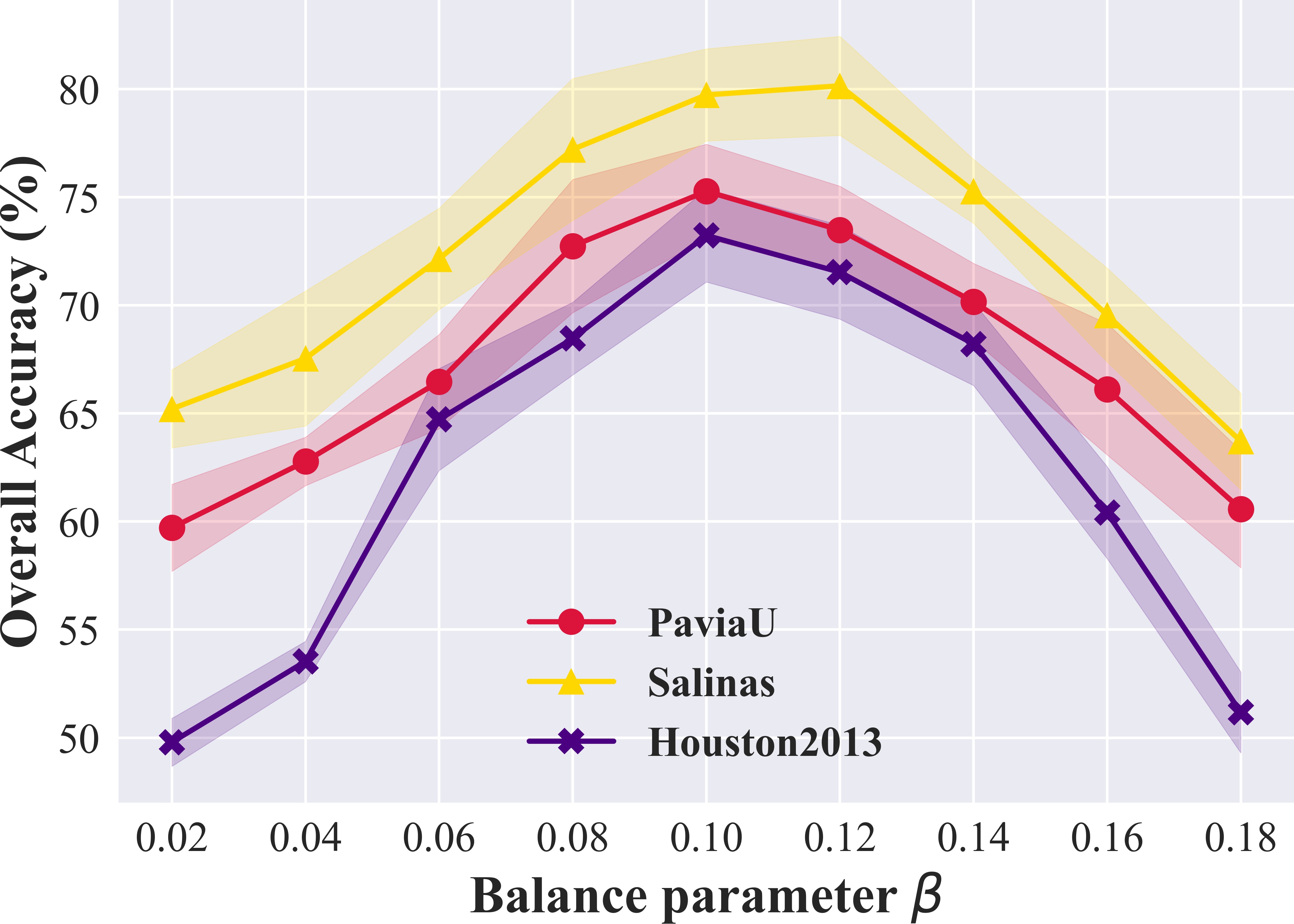}}
   \subfigure[]{\label{fig:d6_b}\includegraphics[width=0.49\linewidth]{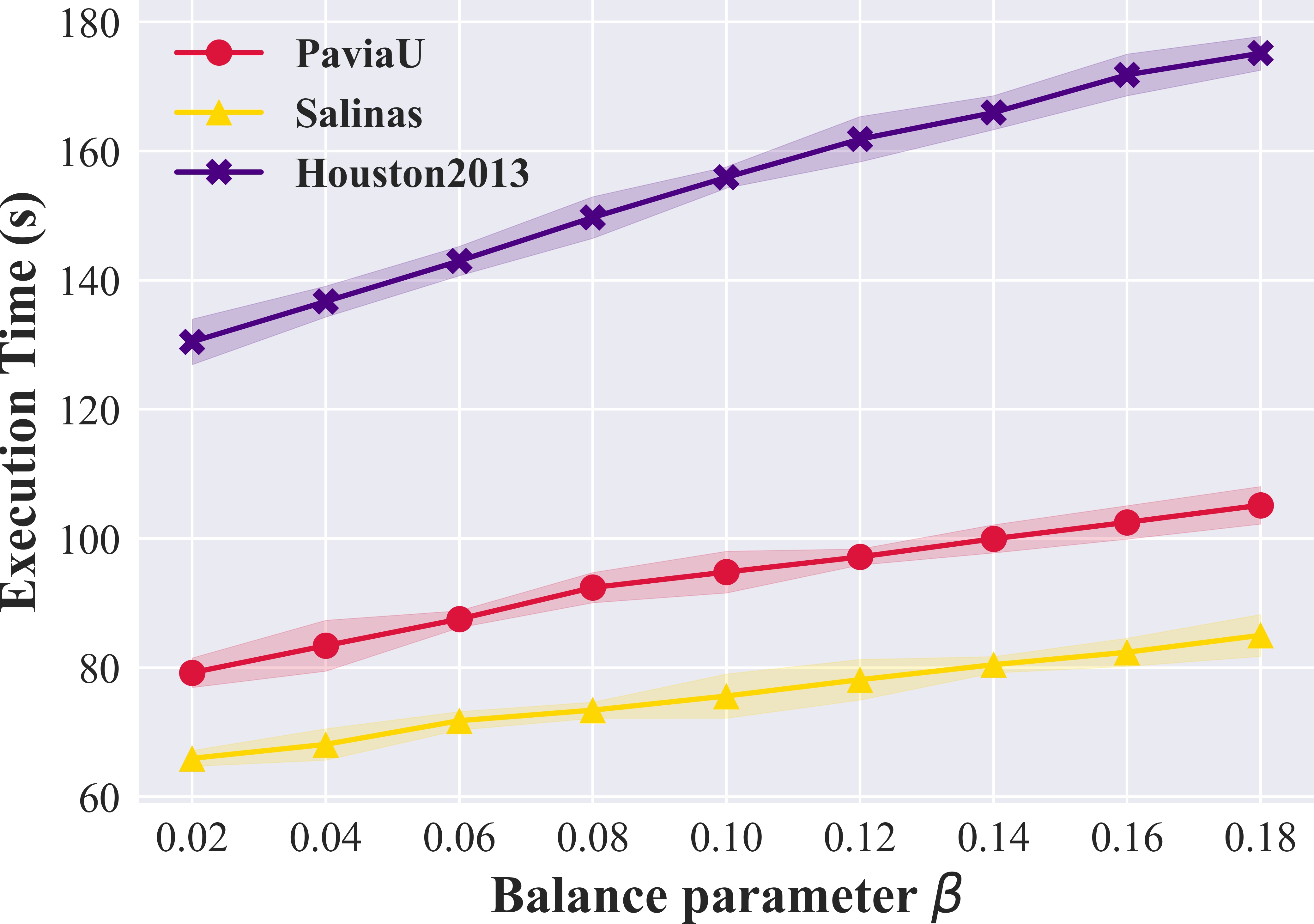}}
     \caption{The hyper-parameters sensitivity analysis results of the balance parameter $\beta$. (a) Effectiveness; (b) Efficiency.}
  \label{fig:D6}
\end{figure}

 
\textbf{2) Hyper-parameters Sensitivity Analysis}: The most important hyper-parameters contained in this work refers to the masking ratio $\alpha$ and balance parameter $\beta$ in dynamic graph structure learning. 
We evaluate the influences of these two hyper-parameters from the effectiveness and efficiency aspects for a comprehensive analysis. 
The evaluation metrics are specified as the average results over different attack methods, which have been illustrated in Fig. \ref{fig:D5} and Fig. \ref{fig:D6}. 
According to Fig. \ref{fig:D5}, it can be inferred that increasing the masking ratio would improve the effectiveness and efficiency of MSSA. 
However, when the masking ratio $\alpha$ is so large that few spectral information is utilized to conduct spectrum reconstruction, MSSA may have a limited performance in this situation.  
\begin{table}[!t]
  \renewcommand{\arraystretch}{1.25}
  \scriptsize
  \begin{center}
  \caption{Quantitative performances of comparing adversarial defense strategies in Salinas dataset.}
\label{table:7}
\begin{tabular}{| c | c | c | c | c | c | c |}
\hline
{Method}     & {Metrics} & {YOPO}  & {TE} & {ALS} & {\cellcolor{lightgray!50} MSSA}\\
\hline\hline
\multirow{3}{*}{FGSM}& {OA(\%)}   &71.77±7.11 & 76.25±6.97 & 44.92±1.84 & \textbf{83.21±1.54}\\
& {AA(\%)}    & 79.32±5.44 & 70.49±0.81 & 54.78±5.7 & \textbf{83.86±5.39} \\ 
& {$\kappa$ (\%)}  & 68.95±4.35 & 73.2±1.42 & 39.02±7.05 & \textbf{81.27±4.43} \\
 \hline\hline
\multirow{3}{*}{BIM}& {OA(\%)}   & 41.12±6.58 & 72.24±0.51 & 44.57±5.38 & \textbf{78.93±4.81} \\ 
& {AA(\%)}   & 50.77±0.58 & 76.71±4.26 & 54.51±4.28 & \textbf{81.57±2.11} \\
& {$\kappa$ (\%)}   & 34.95±3.64 & 72.78±6.63 & 38.6±0.91 & \textbf{76.26±1.18} \\
 \hline\hline
 \multirow{3}{*}{DeepFool}& {OA(\%)}  & 38.69±0.32 & 71.72±4.57 & 23.14±1.71 & \textbf{72.17±1.93} \\
 & {AA(\%)}   & 36.29±0.49 & \textbf{70.19±4} & 33.99±4.02 & 67.73±6.49 \\
 & {$\kappa$ (\%)}  & 32.09±3.09 & 68.47±2.64 & 15.2±7.02 & \textbf{68.56±3.15} \\
\hline
\end{tabular}
\begin{tablenotes}
  \item The \textbf{bold entries} represent the best performance in each row.
\end{tablenotes}
\end{center}
\end{table}
\begin{table}[!t]
  \renewcommand{\arraystretch}{1.25}
  \scriptsize
  \begin{center}
  \caption{Quantitative performances of comparing adversarial defense strategies in Houston2013 dataset.}
\label{table:8}
\begin{tabular}{| c | c | c | c | c | c | c |}
\hline
{Method}     & {Metrics} & {YOPO}  & {TE} & {ALS} & {\cellcolor{lightgray!50} MSSA}\\
\hline\hline
\multirow{3}{*}{FGSM}& {OA(\%)}   & 43.23±7.73 & 40.3±8.46 & 38.29±6.34 & \textbf{79.51±2.93}\\
& {AA(\%)}    & 45.61±3.44 & 33.59±4.88 & 39.85±6.74 & \textbf{80.18±2.53} \\ 
& {$\kappa$ (\%)}  & 38.9±4.12 & 35.07±5.02 & 33.8±7.46 & \textbf{77.81±2.84} \\
 \hline\hline
\multirow{3}{*}{BIM}& {OA(\%)}   & 21.75±3.78 & 39.13±8.9 & 22.57±5.45 & \textbf{71.36±2.93} \\ 
& {AA(\%)}   & 22.48±5.62 & 32.49±8.33 & 26.38±2.8 & \textbf{72.86±2.71} \\
& {$\kappa$ (\%)}   & 15.65±6.57 & 33.85±1.57 & 17.54±8.11 & \textbf{69.05±3.87} \\
 \hline\hline
 \multirow{3}{*}{DeepFool}& {OA(\%)}  & 11.17±6.96 & 37.16±7.06 & 13.54±6.49 & \textbf{70.47±3.76} \\
 & {AA(\%)}   & 13.15±7.07 & 30.28±2.52 & 16.88±5.47 & \textbf{72.03±3.37} \\
 & {$\kappa$ (\%)}  & 4.49±6.68 & 31.67±3.14 & 7.8±5.46 & \textbf{68.03±3.07} \\
\hline
\end{tabular}
\begin{tablenotes}
  \item The \textbf{bold entries} represent the best performance in each row.
\end{tablenotes}
\end{center}
\end{table}

\begin{table}[!t]
  \centering
  \scriptsize
  \renewcommand{\arraystretch}{1.5}
  \begin{threeparttable}
  \caption{THE AVERAGE Execution TIME OF comparing METHODS.} \label{table:9}
  \tabcolsep=4pt
  \setlength{\tabcolsep}{1.49mm}{
  \begin{tabular}{c|cccc}
  \toprule[1pt] 
  \multirow{2}{*}{Data set}    & \multicolumn{4}{c}{Execution time (in seconds) } \\
  \cline{2-5}
                              & Baseline+YOPO & Baseline+TE & Baseline+ALS & \cellcolor{lightgray!50}Baseline+MSSA \\
  \hline
  PaviaU      &124.33±1.92  & 190.43±2.34 & \textbf{57.58±2.96} & 96.81±4.92  \\
  Salinas     &95.58±3.41  & 166.92±3.58 & \textbf{39.98±4.49} & 73.42±2.88  \\
  Houston2013 &192.71±2.96  & 247.15±6.71 & \textbf{92.75±1.51} & 155.93±3.17 \\
  Average     &137.54±2.76  & 201.5±4.21 & \textbf{63.44±2.99} & 108.72±3.66\\
  \bottomrule[1pt]
  \end{tabular}
  \begin{tablenotes}
    \item The \textbf{bold entries} represent the best performance in each row.
  \end{tablenotes}}
  \end{threeparttable}
\end{table}
\begin{table}[!t]
  \centering
  \scriptsize
  \renewcommand{\arraystretch}{1.5}
  \begin{threeparttable}
  \caption{Defense and classification performances of each module in MSSA (Reported in average OA values)} \label{table:10}
  \tabcolsep=4pt
  \setlength{\tabcolsep}{4.1mm}{
  \begin{tabular}{ccc|cc}
  \toprule[1pt] 
   Baseline  &MSAL & DGE & Clean (\%) & Adversarial (\%) \\
  \hline
  \color{red}\checkmark    &     &    &81.33±3.49 &15.19±3.01  \\
  \color{red}\checkmark    &\color{red}\checkmark   &  & 87.91±3.63   &69.59±1.92\\
  \color{red}\checkmark  &     &\color{red}\checkmark      &94.54±2.04 &60.47±3.39\\
  \color{red}\checkmark  & \color{red}\checkmark & \color{red}\checkmark &\textbf{98.14±3.34} & \textbf{76.08±3.26}\\
  \bottomrule[1pt]
  \end{tabular}
  \begin{tablenotes}
    \item The \textbf{bold entries} represent the best performance in each column.
  \end{tablenotes}}
  \end{threeparttable}
\end{table}
Furthermore, it is noticeable that MSSA performs poorly if the masking ratio $\alpha$ is set as a small value, which further validates the necessity of input random masking strategy. 
In terms of the balance parameter $\beta$, if it is set as a proper large value, DGE module could contribute a lot to the final results.
Because a larger $\beta$ is able to introduce more uncertainty to the training process of MSSA. 
But when the value of $\beta$ is too large, substantial uncertainty would be introduced and it is difficult for MSSA to find the global optimal solution. 
\section{conclusion}\label{sec:6}
In this paper, we propose a novel spatial-spectral masked autoencoder (MSSA) to systematically tackle the adversarial attack problem for HSIs. 
MSSA firstly develops a special learning paradigm based on random input masking operation and self-supervised mechanism to improve the stability from spectral aspect. 
On the other hand, the graph theory is exploited for designing a more robust architecture in the spatial aspect by establishing global pixel-wise combinations. 
Besides, we also propose a dynamic graph structure learning scheme to establish a positive and compact feedback between aforementioned two modules. 
Extensive experiments have demonstrated the superiority of MSSA in comparison with the state-of-the-art hyperspectral classifiers and representative defense strategies. 
\par
Furthermore, random input masking, an intuitive and effective technology, has been proven to enhance the robustness of DL models in our research. 
However, we believe that this technology can be generalized to address the adversarial attack issues in other research fields. 
In addition, figuring out how to improve the efficiency of MSSA remains a great challenge and requires further research in the future work.

\ifCLASSOPTIONcaptionsoff
  \newpage
\fi

\end{document}